\documentclass[10pt]{article} % TMLR

\usepackage[accepted]{tmlr}

\usepackage{amsmath,amssymb,amsthm,mathtools}
\usepackage{booktabs}
\usepackage{graphicx}
\usepackage{hyperref}
\usepackage{url}
\hypersetup{pdftitle={DP-Splat: Bayesian Nonparametric Complexity Control for Gaussian Splatting},
            pdfauthor={Aqi Dong}, pdfsubject={Transactions on Machine Learning Research, 2026}}

\newtheorem{proposition}{Proposition}

\newcommand{\Khat}{\texorpdfstring{\widehat{K}}{K-hat}}  % readable in PDF bookmarks
\newcommand{\E}{\mathbb{E}}
\newcommand{\R}{\mathbb{R}}
\newcommand{\Nk}{N_k}
\newcommand{\elbo}{\mathcal{L}}
\newcommand{\NIW}{\mathrm{NIW}}
\newcommand{\Dir}{\mathrm{Dir}}
\newcommand{\Betad}{\mathrm{Beta}}
\newcommand{\Gammad}{\mathrm{Gamma}}
\newcommand{\St}{\mathrm{St}}

\title{DP-Splat: Bayesian Nonparametric Complexity Control\\for Gaussian Splatting}

\author{\name Aqi Dong \email donga2@erau.edu \\
      \addr Embry-Riddle Aeronautical University}

\def\month{09}  % month of publication (camera-ready)
\def\year{2026}
\def\openreview{\url{https://openreview.net/forum?id=75Hx0RDPMr}}

\begin{document}

\maketitle

\begin{abstract}
3D Gaussian Splatting represents scenes as finite mixtures of anisotropic Gaussians whose
number of components $K$ is governed by heuristic adaptive density control or user-specified
caps. Variational Bayes Gaussian Splatting (VBGS) recast splat fitting as conjugate
variational inference over a finite mixture, but $K$ remains fixed. We replace the finite
symmetric Dirichlet over mixture weights with a truncated stick-breaking Dirichlet-process
prior (and, as a theory-backed alternative, a sparse overfitted finite Dirichlet), so that
the number of \emph{occupied} components adapts to the data while every update remains a
closed-form coordinate-ascent step (Beta/Categorical/Normal--Inverse--Wishart); a
natural-gradient stochastic variant makes the per-step cost independent of the number of
points. As in VBGS, the likelihood is a mixture density on colored points: model, theory,
and scene evaluations live in this point space, and rasterized rendering is a downstream
deployment map. We give an ELBO monotonicity-and-convergence guarantee, a rigorous truncation-error bound
that corrects a large-$\alpha$ approximation in common use that undershoots the exact
Ishwaran--James bound it approximates and so cannot be relied on as a surrogate for that
bound, and an
honest account of what the fitted number of components does and does not estimate.
Empirically, (i) the effective complexity $\Khat$ adapts to data complexity and recovers
the true $K$ (within $\pm1$) on well-separated synthetic data with regime-appropriate
concentration, a dependence we map systematically over initializations, restarts, and
concentration settings; (ii) a deconfounded three-way image comparison (all methods
rendered through the reference implementation's shared renderer) shows the DP prior's
contribution is complexity \emph{selection}, not per-component efficiency: converged DP
fits exceed single-pass fixed-$K$ VBGS at matched budgets by $+2.7$\,dB on average yet tie
an equally converged fixed-$K$ baseline, while on all eight NeRF-synthetic scenes, at a
single concentration fixed in advance ($\alpha{=}100$), DP-Splat holds its held-out
point-space color prediction within $0.33$\,dB of VBGS's (ahead on four of the eight,
three of those resolved over three seeds) using
$3.9$--$7.9\times$ fewer occupied components, mean $5.0\times$, with occupancy scored by one
rule for both models,
whereas under full rasterization on strictly held-out views the same model at the same
$\alpha{=}100$ setting renders $3.4$--$9.0$\,dB below the unpruned fixed-$K$ baseline, a
deficit we report rather than tune away and do not explain by the component budget;
(iii) the posterior-predictive color variance is well calibrated on model-matched
synthetic data (regression-ECE $\approx 3\times10^{-3}$), and calibration curves,
central-interval coverage, and held-out predictive log-likelihood are reported across all
eight NeRF-synthetic scenes under misspecification; a seeded matched-capacity ablation
attributes that calibration to the conjugate Normal--Inverse--Wishart color model rather
than to the weight prior, since a fixed-$K$ Dirichlet mixture given the capacity the DP
selects shows no resolved calibration difference (paired differences of $-5\times10^{-5}$ in
regression-ECE and $-9\times10^{-5}$ in coverage-ECE, each a few percent of the seed
spread, with paired 95\% intervals covering zero on all three uncertainty metrics), which leaves automatic complexity selection as the Dirichlet-process prior's
distinct contribution and a small $0.077$\,dB point-PSNR edge to the DP arm as the one
difference the ablation does not resolve; and (iv) the ordering suggested by
exact-posterior asymptotics reverses under mean-field coordinate ascent at practical $N$:
the Dirichlet-process prior \emph{resists} over-splitting while the sparse finite mixture
saturates its truncation, a gap between variational practice and posterior asymptotics
that we document across three orders of magnitude in $N$. Code, exact reproduction
commands, and all experiment records are publicly available.
\end{abstract}

\section{Introduction}
\label{sec:intro}

3D Gaussian Splatting (3DGS) \citep{kerbl2023gaussian} represents a scene as a mixture of
anisotropic 3D Gaussians and renders it with a differentiable rasterizer, achieving
state-of-the-art novel-view synthesis at real-time frame rates. A central but under-examined
design choice is the \emph{number of components}: 3DGS grows and prunes Gaussians with
adaptive density control (gradient-magnitude splitting and opacity thresholds) that works
well in practice but is a stack of heuristics with tunable knobs and no statistical
semantics. Follow-up work has made the dynamics more principled (e.g., MCMC-style
relocation moves; \citealp{kheradmand2024mcmc}) yet still requires a user-specified cap
on the component count.

Variational Bayes Gaussian Splatting (VBGS; \citealp{vandemaele2024vbgs}) reframed
splat fitting as conjugate variational inference over a finite mixture on colored points:
closed-form coordinate-ascent variational inference (CAVI) replaces stochastic gradients, enabling continual
learning without replay. But the mixture is finite with $K$ fixed in advance, uncertainty is
per-component only, and VBGS inherits a component-recycling heuristic to revive dead
components, which is adaptive density control through the back door.

In the statistics literature, letting a mixture choose its own complexity is a mature
problem with three standard routes: Dirichlet-process (DP) mixtures with truncated
variational inference \citep{ferguson1973bayesian,sethuraman1994constructive,%
ishwaran2001gibbs,blei2006variational}; sparse overfitted finite mixtures whose superfluous
components empty asymptotically \citep{rousseau2011asymptotic,malsinerwalli2016model}; and
mixtures of finite mixtures with a prior on $K$ \citep{miller2018mixture}. None of this
machinery has, to our knowledge, been applied to splatting-style scene representations.

\paragraph{This paper.} We replace VBGS's finite symmetric Dirichlet over mixture weights
with a truncated stick-breaking DP prior (primary variant) and a sparse overfitted Dirichlet
(secondary, theory-backed variant), keeping the entire pipeline conjugate: every update
remains a closed-form CAVI step over Beta, Categorical, Normal--Inverse--Wishart (and
optionally Gamma, when the DP concentration $\alpha$ is learned) factors, and a
natural-gradient stochastic variant (SVI; \citealp{hoffman2013stochastic}) has per-step cost
independent of the number of points $N$. The result is a splat-fitting procedure in which
the number of \emph{occupied} components $\Khat$ is an output of posterior inference rather
than an input.

Our contributions are deliberately statistical rather than benchmark-oriented; we validate
claimed properties instead of chasing rendering leaderboards. To keep established
machinery and new material separate, we state the contributions in four groups and tag
the basis each builds on:
\begin{enumerate}
\item \textbf{Integration: closed-form nonparametric splat fitting.} A ``weight-prior
  switch'' ($\mathrm{dp}$ / $\mathrm{sparse\_dir}$ / $\mathrm{dir}$) over an otherwise
  unchanged conjugate model (\S\ref{sec:model}--\S\ref{sec:inference}), with exact CAVI
  (monotone to machine precision; Proposition~\ref{prop:monotone}, a standard argument
  restated) and an SVI variant that reproduces CAVI exactly in the full-batch limit and
  fits $N=10^7$ points on a laptop CPU. The ingredients (truncated stick-breaking VI,
  sparse overfitted mixtures, natural-gradient SVI) are established; their integration
  into splat fitting with full conjugacy preserved is new.
\item \textbf{Empirical diagnosis: what the prior does and does not deliver.} Consistency
  theory says DP posteriors inflate the number of components
  \citep{miller2014inconsistency} while sparse overfitted mixtures empty them
  \citep{rousseau2011asymptotic}. Under CAVI \emph{the ordering reverses}: the DP variant
  shows slow, log-like $\Khat$ growth while the sparse variant saturates its truncation by
  $N=10^6$ regardless of its concentration (Figure~\ref{fig:f4}); both effects trace to
  coordinate-ascent dynamics (component death without revival) rather than to the priors,
  learning $\alpha$ tracks the local optimum instead of correcting it, and a systematic
  sensitivity study over initializations, restarts, and concentration, with ELBO-guarded
  reordering and revival diagnostics, quantifies which conclusions are robust
  (\S\ref{sec:khatsens}). A deconfounded matched-budget comparison completes the picture:
  converged CAVI exceeds single-pass fixed-$K$ VBGS by $+2.7$\,dB on average yet ties an
  equally converged fixed-$K$ baseline at the same $\Khat$, so the prior's contribution is
  complexity selection, not per-component efficiency and not the predictive
  uncertainty, which contribution~4 below traces to the color model instead; across all
  eight
  NeRF-synthetic scenes, at the single concentration $\alpha{=}100$ fixed in advance,
  DP-Splat's held-out point-space color prediction (scored on a single held-out frame
  per scene under the reference implementation's protocol, in contrast to the 25-view
  rasterized evaluation of \S\ref{sec:raster}) stays within $0.33$\,dB of VBGS's and is
  ahead on four of the eight (three of them resolved over three seeds), using
  $3.9$--$7.9\times$ fewer effective components
  (mean $5.0\times$), with occupancy scored for both models by the same $\Nk > 1$ rule on a
  full-data E-step (\S\ref{sec:experiments}). $\Khat$ is a
  useful, pragmatic complexity control; it is not a consistent estimator of a ``true''
  $K$, and we say so.
\item \textbf{Correction: a rigorous truncation-error bound.} We state the closed-form
  finite-$\alpha$ truncation bound $2N(\alpha/(1{+}\alpha))^{T-1}$ (new) and show that the
  widely quoted $4Ne^{-(T-1)/\alpha}$ approximation of \citet{ishwaran2001gibbs} is
  \emph{anti-conservative} at $\alpha\approx 1$: it undershoots even the exactly evaluated
  Ishwaran--James bound by orders of magnitude in the regime where DP-Splat models
  actually operate, and so cannot be relied on as a surrogate for the bound it
  approximates (\S\ref{sec:theory}, Figure~\ref{fig:f3}).
\item \textbf{Calibrated color uncertainty, and an ablation that locates its source.} The
  mixture-of-Students posterior
  predictive (standard NIW machinery) yields a closed-form conditional color variance
  $\mathrm{Var}[c\,|\,s]$ whose binned calibration error against realized squared error is
  $\approx 3\times10^{-3}$ on model-matched synthetic data (Figure~\ref{fig:f5}). The
  evaluation is point-space throughout: calibration curves, central-interval coverage,
  and held-out predictive log-likelihood on all eight NeRF-synthetic scenes, reported
  under misspecification (\S\ref{sec:uncertainty8}); in fixed-precision baselines
  the same quantity is constant and non-adaptive. We do not credit this to the
  Dirichlet-process prior. A seeded matched-capacity ablation, in which a fixed-$K$
  symmetric Dirichlet mixture is given the capacity the DP selects and run under an
  identical protocol, shows no resolved calibration difference from the DP fit on any of
  the three uncertainty metrics: the paired differences over the eight scenes are
  $-5\times10^{-5}$ in regression-ECE (95\% interval $[-7.2\times10^{-4},
  +6.1\times10^{-4}]$), $-9\times10^{-5}$ in coverage-ECE ($[-1.6\times10^{-3},
  +1.4\times10^{-3}]$) and $-0.010$ nats in conditional held-out log-likelihood
  ($[-0.038, +0.019]$), every interval covering zero. Calibrated predictive color uncertainty is a property of the conjugate NIW
  color model, and what the DP prior contributes that a fixed-$K$ mixture cannot is
  choosing the capacity in the first place. The two priors are not identical in every
  respect: a $0.077$\,dB point-PSNR edge to the DP arm survives the ablation without being
  resolved by it (\S\ref{sec:uncertainty8}).
\end{enumerate}

Throughout, the likelihood is the mixture density on colored points (as in VBGS): rendering
is a deployment map applied to the fitted posterior, not part of the model, and all theory
is stated for the point mixture (\S\ref{sec:theory}). Image comparisons
(\S\ref{sec:experiments}) render every method through the reference implementation's
shared expected-color renderer; scene-level claims are point-space, and rasterized
novel-view evaluation is reported separately (\S\ref{sec:raster}).

\section{Related work}
\label{sec:related}

\paragraph{Complexity control in splatting.} 3DGS \citep{kerbl2023gaussian} manages its
component budget by gradient-magnitude densification and opacity pruning, effective but
untheorized heuristics whose thresholds are tuned per scene family. 3DGS-as-MCMC
\citep{kheradmand2024mcmc} replaces cloning/splitting with probability-preserving
relocation moves, giving the dynamics a sampling interpretation, but retains a
user-specified cap on the number of Gaussians. BOGausS \citep{pateux2025bogauss} attaches
variational-flavored confidence scores to densification decisions. VBGS
\citep{vandemaele2024vbgs} is our direct foundation: it fits the mixture on colored
point data with conjugate priors and closed-form CAVI, enabling continual learning, and has
been extended to SLAM \citep{zhu2026vbgsslam}; in all of these the mixture is finite with $K$
fixed, and VBGS additionally relies on a reassignment heuristic that recycles dead
components. A separate engineering line compacts \emph{trained} 3DGS models (post-hoc
importance pruning, \citealp{fan2024lightgaussian}; learned masks, \citealp{lee2024compact};
Hessian-based uncertainty pruning, \citealp{hanson2025pup}), demonstrating that standard
fits are heavily over-parameterized; these prune a fixed-budget MLE fit after the fact,
whereas we build parsimony into the prior and obtain it during inference, with an account
of what the resulting count means. Our work replaces exactly one ingredient of VBGS, the
weight prior, and removes the need to choose $K$; notably, VBGS's own default
concentration ($e_0 = 1/K$) already makes its finite Dirichlet ``sparse'' in the sense of
\citet{rousseau2011asymptotic}, a fact we make explicit and exploit as a baseline variant.

\paragraph{Automatic complexity in mixtures.} The three standard statistical routes are:
(i) DP mixtures \citep{ferguson1973bayesian,sethuraman1994constructive} with blocked-Gibbs
or truncated variational inference \citep{ishwaran2001gibbs,blei2006variational};
(ii) sparse overfitted finite mixtures, where a concentration below a dimension-dependent
threshold empties superfluous components asymptotically
\citep{rousseau2011asymptotic,malsinerwalli2016model}; and (iii) mixtures of finite
mixtures with a prior on $K$, the route with consistent posterior inference on the number of
components \citep{miller2018mixture,fruhwirthschnatter2021generalized}. The cautionary
results of \citet{miller2013simple,miller2014inconsistency}, namely that DP and Pitman--Yor
mixture posteriors are inconsistent for the number of components, anchor our discussion of
what $\Khat$ estimates. Posterior contraction of the mixing measure in Wasserstein distance is
due to \citet{nguyen2013convergence}. Our contribution to this literature is not new
asymptotics but a documented case study of how \emph{variational} fitting dynamics interact
with (and at practical sample sizes, dominate) these prior-level asymptotics.

\paragraph{Variational methods.} We use textbook conjugate mean-field machinery
\citep{bishop2006pattern,blei2017variational}, stochastic natural-gradient updates
\citep{hoffman2013stochastic,amari1998natural}, and the standard NIW posterior-predictive
Student-$t$ \citep{murphy2012machine}; component emptying under variational Bayes for
mixtures was observed as early as \citet{ghahramani2000variational}. Closed-form
coordinate ascent in models with Student-$t$ structure is an active topic, most recently
for Student-$t$ process regression via scale-mixture augmentation
\citep{onoue2026closedform}; a parallel line makes heavy-tailed models scalable by
building sparse inducing-point Student-$t$ \emph{processes}, first with a variational
lower bound for the process itself \citep{xu2024sparsevstp} and subsequently with an
extended treatment of heavy-tailed regression \citep{xu2026sparsevstpnn}. In all three
the Student-$t$ is the modeling commitment, a heavy-tailed class posited a priori and
equipped with its own variational machinery; in our setting the Student-$t$ is not a
modeling choice but the analytic NIW posterior predictive of our finite conjugate
mixture, so the tails and the per-point predictive variance follow from marginalizing
the color mean and covariance under the standard conjugate mean-field update. We cite
this line to position our predictive, not to adopt its inference. The order-dependence
of the truncated stick-breaking family and size-based component reordering were identified
by \citet{kurihara2007collapsed}; we deliberately implement the plain scheme and quantify
its failure modes instead of patching them (\S\ref{sec:experiments},
\S\ref{sec:khatsens}).

\paragraph{Uncertainty for scene representations.} Post-hoc or architectural uncertainty
for radiance fields and splatting is an active area (e.g., Fisher-information view
selection; \citealp{jiang2024fisherrf}); VBGS exposes per-component parameter uncertainty
but fixes its color precision, so its predictive color variance is constant and
non-adaptive rather than inferred per point. Our mixture-of-Students
predictive gives calibrated per-point color variance as a by-product of the posterior
(\S\ref{sec:experiments}), which we validate with regression-style calibration curves
\citep{kuleshov2018accurate}.

\section{Model}
\label{sec:model}

\paragraph{Data model.} Data are $N$ colored points; in 2D-image mode each pixel is a point.
Point $n$ has a spatial coordinate $s_n \in \R^{D_s}$ ($D_s{=}3$ for scenes, $2$ for images)
and a color $c_n \in \R^{D_c}$ ($D_c{=}3$). Write $x_n = (s_n, c_n)$. Conditional on the
assignment $z_n = k$ the two modalities are independent Gaussians,
\begin{equation}
s_n \mid z_n{=}k \sim \mathcal{N}(\mu_{k,s}, \Sigma_{k,s}), \qquad
c_n \mid z_n{=}k \sim \mathcal{N}(\mu_{k,c}, \Sigma_{k,c}),
\label{eq:datamodel}
\end{equation}
as in VBGS, whose likelihood we inherit unchanged except that we place a full conjugate
prior on the color covariance where VBGS fixes its color precision (we retain its fixed
variant behind a compatibility flag used for cross-code regression tests,
\S\ref{sec:experiments}). The likelihood is the mixture density on points obtained from
RGB-D or lifted depth; rendering (expected color conditional on projected location, with
the standard rasterizer handling occlusion) is a deployment map, not part of the
likelihood. The color factor in Eq.~\eqref{eq:datamodel} is a single Gaussian per
component with no spherical-harmonic or viewing-direction term, so the model class is
view-independent; we return to this limitation in \S\ref{sec:discussion}.

\paragraph{Component priors.} Per modality $m \in \{s, c\}$, conjugate
Normal--Inverse--Wishart:
\begin{equation}
\Sigma_{k,m} \sim \mathcal{IW}(\Psi_{0,m}, \nu_{0,m}), \qquad
\mu_{k,m} \mid \Sigma_{k,m} \sim \mathcal{N}\big(m_{0,m},\, \Sigma_{k,m}/\kappa_{0,m}\big),
\label{eq:niwprior}
\end{equation}
with defaults $m_{0,m}$ the data centroid, $\Psi_{0,m} = \widehat{\sigma}_m^2 I
(\nu_0 - D_m - 1)$ with $\widehat{\sigma}_m^2$ the mean per-dimension variance,
$\kappa_0 = 10^{-3}$, $\nu_0 = D_m + 2$.

\paragraph{Weight priors (the switch).} \emph{Variant A (primary), truncated stick-breaking
DP:}
\begin{equation}
v_k \sim \Betad(1, \alpha)\ (k < T), \qquad v_T \coloneqq 1, \qquad
\pi_k = v_k \textstyle\prod_{j<k}(1 - v_j),
\label{eq:stick}
\end{equation}
where $T$ is an inference-time truncation (the Gaussian budget for scenes), optionally with
a conjugate $\alpha \sim \Gammad(a_0, b_0)$ ($a_0 = b_0 = 1$; flag-controlled).
\emph{Variant B (secondary), sparse overfitted finite mixture:}
$\pi \sim \Dir(e_0, \dots, e_0)$ with $K = T$ components and $e_0 \ll 1$; by
\citet{rousseau2011asymptotic}, $e_0 < d/2$ (with $d$ the per-component parameter dimension)
makes superfluous components empty asymptotically. \emph{Baseline:} the same Dirichlet with
VBGS's shipped $e_0 = 1/K$. All three run through one switch on an otherwise identical
model. Each of these weight priors and every conjugate update in \S\ref{sec:inference}
is established machinery; the only new element of the model is the switch itself, which
swaps the weight prior while leaving the VBGS point likelihood, the NIW component priors,
and the mean-field family unchanged.

\paragraph{Variational family.} Truncated mean field,
\begin{equation}
q(z, v, \theta, \alpha) \;=\; \prod_{n=1}^{N} q(z_n)\,
\prod_{k=1}^{T-1} q(v_k)\, \prod_{k=1}^{T} \prod_{m} q(\mu_{k,m}, \Sigma_{k,m})\; q(\alpha),
\label{eq:meanfield}
\end{equation}
with $q(z_n) = \mathrm{Cat}(r_{n,1:T})$, $q(v_k) = \Betad(\gamma_{k,1}, \gamma_{k,2})$,
$q(\mu, \Sigma) = \NIW(m_k, \kappa_k, \Psi_k, \nu_k)$ per modality, and
$q(\alpha) = \Gammad(w_1, w_2)$ when $\alpha$ is learned.

\section{Inference}
\label{sec:inference}

\paragraph{CAVI.} With soft counts and weighted moments
$\Nk = \sum_n r_{nk}$, $\bar{x}_k = \Nk^{-1}\sum_n r_{nk} x_n$,
$S_k = \sum_n r_{nk}(x_n - \bar{x}_k)(x_n - \bar{x}_k)^\top$ (per modality), coordinate
ascent cycles four exact updates until the relative ELBO change is below $10^{-6}$:

\emph{(i) NIW (each modality independently):}
\begin{equation}
\kappa_k = \kappa_0 + \Nk,\quad
m_k = \frac{\kappa_0 m_0 + \Nk \bar{x}_k}{\kappa_k},\quad
\nu_k = \nu_0 + \Nk,\quad
\Psi_k = \Psi_0 + S_k + \tfrac{\kappa_0 \Nk}{\kappa_0 + \Nk}
(\bar{x}_k - m_0)(\bar{x}_k - m_0)^\top.
\label{eq:niwupdate}
\end{equation}

\emph{(ii) Sticks (Variant A):}
$\gamma_{k,1} = 1 + \Nk$, $\gamma_{k,2} = \E_q[\alpha] + \sum_{j>k} N_j$; the Dirichlet
variants use $\alpha^{\mathrm{post}}_k = e_0 + \Nk$.

\emph{(iii) Responsibilities} via expected log-weights
$\E[\log \pi_k] = \E[\log v_k] + \sum_{j<k}\E[\log(1 - v_j)]$ (digamma identities;
Dirichlet form $\psi(\alpha_k^{\mathrm{post}}) - \psi(\sum_j \alpha_j^{\mathrm{post}})$) and
the expected Gaussian log-density under the NIW posterior
\citep[\S10.2]{bishop2006pattern}:
\begin{equation}
\log \rho_{nk} = \E[\log \pi_k] + \sum_{m}\Big(
\tfrac{1}{2}\E[\log|\Lambda_{k,m}|] - \tfrac{D_m}{2}\log 2\pi
- \tfrac{1}{2}\E\big[(x_{n,m} - \mu_{k,m})^\top \Lambda_{k,m}(x_{n,m} - \mu_{k,m})\big]
\Big),
\label{eq:resp}
\end{equation}
normalized with log-sum-exp.

\emph{(iv) Concentration (flagged):} $w_1 = a_0 + T - 1$,
$w_2 = b_0 - \sum_{k<T}\E[\log(1 - v_k)]$.

The ELBO is implemented one named function per term and each expectation is verified against
Monte Carlo estimates from $q$; the full loop is verified against an independently written
brute-force oracle and against the reference VBGS implementation (\S\ref{sec:experiments}).
Means are initialized by $k$-means++ seeding on a subsample; all other posteriors start at
their priors.

\paragraph{Stochastic variant.} For large $N$, the global parameters
$\lambda \in \{\gamma, \text{NIW naturals}, (w_1, w_2)\}$ follow natural-gradient steps
\citep{hoffman2013stochastic}: a minibatch $B$ yields intermediate estimates
$\hat{\lambda}$ from sufficient statistics scaled by $N/|B|$, blended as
$\lambda^{(t)} = (1 - \rho_t)\lambda^{(t-1)} + \rho_t \hat{\lambda}$ with
$\rho_t = (t + \tau_0)^{-\kappa_{\mathrm{sched}}}$ ($\tau_0 = 64$; $\kappa_{\mathrm{sched}} = 0.7$, distinct from the NIW parameter $\kappa_k$; $|B| = 2^{16}$). The NIW
blend is performed in natural-parameter space
$(\kappa,\ \kappa m,\ \nu,\ \Psi + \kappa m m^\top)$, in which the conjugate update is
affine in the sufficient statistics; with $|B| = N$ and $\rho_t = 1$ the scheme reproduces
full-batch CAVI to $10^{-8}$--$10^{-9}$ relative (unit-tested per parameter block). Per-step cost is independent of $N$: fitting
$N = 10^7$ points takes 12\,s on a laptop CPU (\S\ref{sec:experiments}). The rank-one
subtraction recovering $\Psi$ from naturals requires float64 or centered data; we document
this and all numerical safeguards in the repository.

\paragraph{Effective complexity and rendering.} We report
$\Khat = \#\{k : \Nk > n_{\min}\}$ with $n_{\min} = 1$ (sensitivity to
$n_{\min} \in \{0.5, 1, 2, 5\}$ reported throughout; it never changes a conclusion) and the
entropy-based count $\exp(-\sum_k \tilde{\pi}_k \log \tilde{\pi}_k)$, with
$\tilde{\pi}_k$ the normalized expected weights. For rendering we map
the posterior exactly as VBGS does: component weights $\E_q[\pi_k]$ (posterior expected
stick weights for Variant A), splat shape from $\E[\Sigma_{k}] = \Psi_k/(\nu_k - D - 1)$,
components with $\Nk \le n_{\min}$ dropped; images use VBGS's pure-JAX expected-color
renderer, so image-quality comparisons isolate the weight prior rather than the renderer.

\paragraph{Predictive distribution.} Integrating $(\mu, \Sigma)$ over the NIW posterior
gives a per-component Student-$t$,
$p(x^\ast | k) = \St\!\big(m_k,\ \Psi_k \tfrac{\kappa_k + 1}{\kappa_k \eta_k},\ \eta_k\big)$
with $\eta_k = \nu_k - D + 1$ \citep{murphy2012machine}, and mean-field independence makes
the predictive an exact mixture $\hat{p}(x^\ast) = \sum_k \E[\pi_k]\,
\St_s(s^\ast|k)\,\St_c(c^\ast|k)$. Conditioning on location gives closed-form color moments
\begin{equation}
w_k(s^\ast) \propto \E[\pi_k]\, \St_s(s^\ast \mid k), \quad
\E[c\,|\,s^\ast] = \sum_k w_k m_{k,c}, \quad
\mathrm{Cov}[c\,|\,s^\ast] = \sum_k w_k \big(C_k + m_{k,c} m_{k,c}^\top\big)
- \E[c\,|\,s^\ast]\E[c\,|\,s^\ast]^\top,
\label{eq:condmoments}
\end{equation}
with $C_k = \tilde{S}_{k,c}\,\eta_{k,c}/(\eta_{k,c}-2)$ the Student-$t$ covariance, where
$\tilde{S}_{k,c} = \Psi_{k,c}\,\tfrac{\kappa_{k}+1}{\kappa_{k}\eta_{k,c}}$ is the color
Student-$t$ scale (distinct from the scatter matrix $S_k$; finite covariance for
$\eta_{k,c} > 2$, which also covers existence of the mean, $\eta_{k,c} > 1$; under the
default prior $\eta_{k,c} = \Nk + 3$, so the condition always holds.) All expectations are
with respect to $q$; the factorized per-component predictive
$\St_s \times \St_c$ reflects mean-field independence of the spatial and color blocks. A
complete derivation is given in Appendix~\ref{app:eq7}; the closed form is additionally
verified against independent numerical integration of the joint predictive (agreement
$\le 10^{-11}$) and the implementation assembles the covariance in the
positive-semidefinite law-of-total-variance form. These moments feed the held-out
log-likelihood and calibration experiments.

\section{Statistical properties}
\label{sec:theory}

All statements are for the point-data mixture likelihood \eqref{eq:datamodel}; the
rasterized rendering map is downstream of inference and carries no guarantees here
(\S\ref{sec:discussion}).

\paragraph{What is established and what is new.} To keep attribution unambiguous:
Proposition~\ref{prop:monotone} is the standard coordinate-ascent argument for conjugate
mean-field families \citep[\S10.2]{bishop2006pattern}, see also
\citet{blei2017variational}, restated with
explicit hypotheses for this model and claiming no more than that argument delivers; the
first inequality of Proposition~\ref{prop:truncation} is due to \citet{ishwaran2001gibbs};
the inconsistency and emptying results quoted below are due to
\citet{miller2013simple,miller2014inconsistency} and \citet{rousseau2011asymptotic}. New
in this section are the closed-form finite-$\alpha$ bound (the second inequality of
Proposition~\ref{prop:truncation}) and the analysis showing the popular exponential
approximation to be anti-conservative in the operating regime.

\begin{proposition}[ELBO monotonicity and convergence; standard argument, restated]
\label{prop:monotone}
Assume valid hyperparameters, namely for each modality $m$: $\nu_{0,m} > D_m + 1$,
$\Psi_{0,m}$ positive definite, $\kappa_{0,m} > 0$, and (when $\alpha$ is learned)
$a_0, b_0 > 0$, so that every block update is well defined and the ELBO is bounded above
by the finite log marginal likelihood. Then:
(i) each of the four updates of \S\ref{sec:inference} exactly maximizes the ELBO over its
mean-field block given the others, so the ELBO sequence produced by the CAVI cycle is
nondecreasing and converges to a finite limit;
(ii) for a fixed finite data set, every accumulation point of the iterate sequence is a
coordinate-wise maximizer of the ELBO, that is, a fixed point of the CAVI map.
\end{proposition}

\emph{Argument.} The model is conjugate exponential-family in every block (Categorical,
Beta, NIW, Gamma), so each coordinate update sets the block to the exact conditional
maximizer \citep[\S10.2]{bishop2006pattern}; (i) is coordinate ascent on a function
bounded above. For (ii) the hyperparameter assumptions already do the work, and no
extra regularity condition on the soft counts is needed. By \eqref{eq:niwupdate},
$\kappa_{k,m} = \kappa_{0,m} + \Nk \ge \kappa_{0,m} > 0$,
$\nu_{k,m} = \nu_{0,m} + \Nk > D_m + 1$ and $\Psi_{k,m} \succeq \Psi_{0,m} \succ 0$ hold
for \emph{every} responsibility matrix, so the NIW posteriors are proper and their
log-normalizers smooth however the soft counts behave; a hypothesis phrased as ``$\Nk$
bounded away from configurations that degenerate the NIW posteriors'' would therefore be
vacuous under this prior. What is actually required is boundedness of the iterates, and
that follows from the data. The location parameter
$m_{k,m} = (\kappa_{0,m}/\kappa_{k,m})\, m_{0,m} + (\Nk/\kappa_{k,m})\, \bar{x}_{k,m}$ is
a convex combination, so $\kappa_{0,m} > 0$ and a finite (hence bounded) sample confine it
to the convex hull of $\{m_{0,m}\} \cup \{x_{n,m}\}_{n \le N}$; consequently
$\Psi_{0,m} \preceq \Psi_{k,m} \preceq \Psi_{0,m} + c N I$ with $c$ fixed by the diameter
of that hull, $\Nk \in [0, N]$ bounds $\kappa_{k,m}$, $\nu_{k,m}$ and
$\gamma_{k,1} = 1 + \Nk$, and $\E[\log(1 - v_k)] \le 0$ gives $w_2 \ge b_0 > 0$, hence
$\E_q[\alpha] \le (a_0 + T - 1)/b_0$ and a bound on $\gamma_{k,2}$; the responsibilities
lie in a product of simplices. The iterates thus stay in a compact set on which each
block maximizer is unique (each update minimizes a strictly convex Kullback--Leibler
objective over its block) and depends continuously on the remaining blocks. Every
subsequence then has a convergent sub-subsequence; along it the ELBO values converge to
the finite limit of (i), so by continuity a further block update at the limit point
cannot increase the ELBO, and the limit point already maximizes every block. This is the
standard block-coordinate conclusion \citep[Prop.~2.7.1]{bertsekas1999nonlinear}.
We claim no convergence of the iterates themselves, no uniqueness of the
limit, and no global optimality; the local optima reported in \S\ref{sec:experiments} are
consistent with this. The numerical companion: across all Figure~\ref{fig:f1} runs (20
seeds $\times$ 4 variants, up to 200 iterations each) the worst observed relative ELBO
decrease is $-9.7\times10^{-16}$, a few float64 ulps of the ELBO magnitude, and a separate
unit test runs the full 200 iterations $\times$ 20 seeds $\times$ 4 variants with no
violation beyond $10^{-10}$ relative.

\begin{proposition}[Truncation error, rigorous finite-$\alpha$ form]
\label{prop:truncation}
Let $m_T$ and $m_\infty$ denote the marginal densities of $N$ observations under the
truncated ($v_T \coloneqq 1$) and full stick-breaking priors with concentration $\alpha$,
all else equal. Then
\begin{equation}
\lVert m_T - m_\infty \rVert_1
\;\le\; 2\,\Big(1 - \E\Big[\big(\textstyle\sum_{k<T}\pi_k\big)^{N}\Big]\Big)
\;\le\; 2N\,\Big(\frac{\alpha}{1+\alpha}\Big)^{T-1}.
\label{eq:truncbound}
\end{equation}
\end{proposition}

\emph{Argument.} The first inequality is due to \citet{ishwaran2001gibbs}. For the second,
$1 - x^N \le N(1 - x)$ on $[0,1]$ and
$\E[1 - \sum_{k<T}\pi_k] = \E[\prod_{k<T}(1 - v_k)] = (\alpha/(1{+}\alpha))^{T-1}$ by
independence of the $v_k$. The relation to
\citeauthor{ishwaran2001gibbs}'s widely quoted approximation $4Ne^{-(T-1)/\alpha}$ is as
follows: since $\log(1 + 1/\alpha) \le 1/\alpha$, the per-stick factor obeys
$e^{-1/\alpha} \le \alpha/(1{+}\alpha)$ for \emph{every} $\alpha > 0$ (with near-equality
as $\alpha \to \infty$, the gap being $O(\alpha^{-2})$), so the exponential form decays
strictly faster in $T$ than the rigorous factor; its constant-4 prefactor compensates only
for small $T$ (at $\alpha = 1$, $4Ne^{-(T-1)}$ falls below
$2N\,2^{-(T-1)}$ for all $T \ge 4$). Consequently the exponential expression is
\emph{anti-conservative} in the small-$\alpha$ regime where splat models operate: at
$\alpha \approx 1$ it undershoots even the exactly evaluated intermediate bound
$2(1 - \E[(\sum_{k<T}\pi_k)^N])$ by 3--6 orders of magnitude over the $T$ we sweep
(Figure~\ref{fig:f3}, right), and it should not be quoted as a bound there.
Figure~\ref{fig:f3} overlays that exactly evaluated intermediate bound (by Monte Carlo),
the closed-form bound \eqref{eq:truncbound}, and the exponential approximation.
Practically, at scene budgets ($T$ in the hundreds to thousands, $\Khat \ll T$) the bound
is astronomically small: truncation is a non-issue, and we observe held-out log-likelihood
and $\Khat$ to be flat in $T$ for $T \gtrsim 2.5\,\Khat$.

\paragraph{What $\Khat$ estimates (and does not).} Even exactly computed DP-mixture
posteriors are inconsistent for the number of components: superfluous small clusters persist
as $N \to \infty$ \citep{miller2013simple,miller2014inconsistency}. Sparse overfitted
finite mixtures have the opposite asymptotics, superfluous components empty
\citep{rousseau2011asymptotic}, and mixtures of finite mixtures give the consistent route
\citep{miller2018mixture,fruhwirthschnatter2021generalized}. Our position: $\Khat$ under
truncated CAVI is a \emph{pragmatic complexity control}, not an estimator of a ``true'' $K$
(ill-defined for real scenes in any case), and \S\ref{sec:experiments} shows that at
practical $N$ the fitting dynamics dominate these prior-level asymptotics, to the point of
\emph{inverting} the expected ordering between the DP and sparse variants. We view
quantifying this gap between variational practice and posterior asymptotics as a
contribution in itself; closing it (merge/split or reordering moves,
\citealp{kurihara2007collapsed}; or MFM-style inference) is future work, and the
diagnostic reordering and revival arms of \S\ref{sec:khatsens} give that program a
measured baseline.

\paragraph{Contraction of the mixing measure.} For the fitted mixing measure
$\widehat{G} = \sum_k \E[\pi_k]\,\delta_{(m_{k,s}, m_{k,c})}$ we report Wasserstein-2
distance to the true mixing measure on synthetic data (Figure~\ref{fig:f5}, right);
\citet{nguyen2013convergence} gives posterior contraction rates for mixing measures in
Wasserstein distance for exact posteriors, which we cite as context; no claim is made that
CAVI attains them.

\section{Experiments}
\label{sec:experiments}

Experiments validate claimed statistical properties; we do not tune for rendering
benchmarks. All runs are config-driven scripts with fixed seeds, every reported number is
written to a machine-readable record by the producing script, and all results except
\S\ref{sec:raster} were produced on a single laptop-class CPU (Apple M-series, float64);
the stochastic variant makes this sufficient up to $N = 10^7$.

\paragraph{Verification protocol.} Three independent layers guard correctness: (i) every
update equation is tested against a brute-force NumPy oracle written directly from the
model equations ($10^{-7}$--$10^{-12}$ relative, parameter-dependent); (ii) every ELBO
term is tested against Monte Carlo estimates from $q$; (iii) a cross-code regression
verifies that one step of our CAVI in the baseline configuration ($e_0 = 1/K$, fixed
color precision) reproduces one step of the reference VBGS \texttt{fit\_gmm} to
$10^{-6}$--$10^{-8}$ relative (Appendix~\ref{app:elbo},
Table~\ref{tab:acceptance}). The suite comprises 47 tests.

\subsection{Exactness and convergence (Figure~\ref{fig:f1}, Appendix~\ref{app:elbo})}
Across 20 seeds $\times$ 4 variants ($\mathrm{dp}$, $\mathrm{dp(learn)}$,
$\mathrm{sparse\_dir}$, $\mathrm{dir}$) on colored 2D mixtures ($K_{\mathrm{true}}{=}10$,
$N{=}10^4$, $T{=}30$), the ELBO is monotone to machine precision (worst relative decrease
$-9.7\times10^{-16}$, on one seed of one variant; the other three variants show exactly
zero violations). DP variants converge in 5--16 iterations versus 30--73 (median ${\approx}38$)
for the Dirichlet variants. $2$ of the $20$ DP seeds ($10\%$; $95\%$ Wilson interval
$3$--$30\%$) settle in a visibly lower local optimum, final ELBO ${\approx}6.1\times10^3$
against ${\approx}1.1\times10^4$ for the other 18; CAVI is a local method and we report
this rather than reseeding it away (\S\ref{sec:khatsens} quantifies the restart economics).

\begin{figure}[t]
\centering
\includegraphics[width=0.8\linewidth]{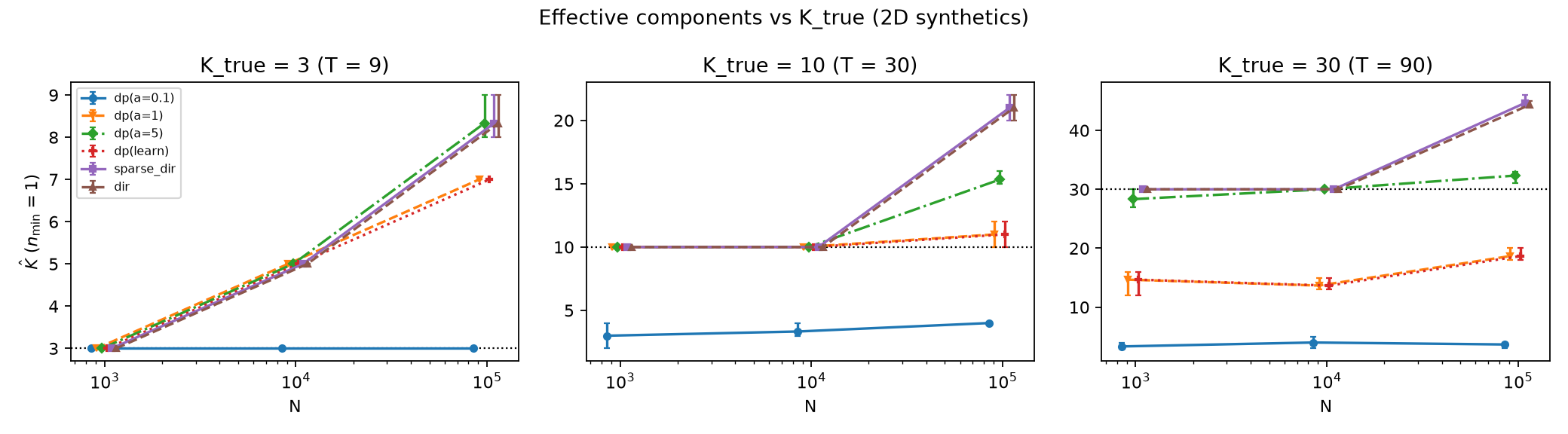}
\caption{$\Khat$ ($n_{\min}{=}1$) vs.\ $N$ for $K_{\mathrm{true}} \in \{3, 10, 30\}$
(dotted line), $T = 3K_{\mathrm{true}}$; bars: seed min/max ($n{=}3$), series offset for
visibility. $\mathrm{dp(learn)}$ coincides with $\mathrm{dp}(\alpha{=}1)$ and
$\mathrm{sparse\_dir}$ with $\mathrm{dir}$, a finding in itself (see text).}
\label{fig:f2}
\end{figure}

\subsection{$\Khat$ recovery is real but concentration-regime-dependent
(Figure~\ref{fig:f2})}
On well-separated colored 2D mixtures with $K_{\mathrm{true}} \in \{3, 10, 30\}$,
$N \in \{10^3, 10^4, 10^5\}$, $T = 3K_{\mathrm{true}}$: with a regime-appropriate fixed
$\alpha$, the DP variant recovers $K_{\mathrm{true}}$ within $\pm 1$, and often
exactly, in every regime ($\alpha{=}0.1 \leftrightarrow K_{\mathrm{true}}{=}3$, exact at
all $N$ including $10^5$; $\alpha{=}1 \leftrightarrow 10$; $\alpha{=}5 \leftrightarrow 30$
at $N{=}10^4$). No single fixed $\alpha$ covers all regimes; small $\alpha$ over-merges
large-$K$ data and large $\alpha$ over-splits small-$K$ data at large $N$. Learning
$\alpha$ variationally does \emph{not} escape this: $\mathrm{dp(learn)}$ tracks
$\mathrm{dp}(\alpha{=}1)$ cell-for-cell; the $\alpha$-posterior adapts to the local optimum
the fit is already in (mode-following), a known but rarely quantified VB pathology, and
Figure~\ref{fig:alphatraj} traces the mechanism cycle by cycle. The
threshold $n_{\min}$ is immaterial: $\Khat$ is identical across
$n_{\min} \in \{0.5, 1, 2, 5\}$ in 159 of 162 fits and differs by at most 1 in the
remainder. The components driving over- or under-estimation carry substantial mass (they
are not threshold artifacts) and no cell's conclusion changes.

\begin{figure}[t]
\centering
\includegraphics[width=0.94\linewidth]{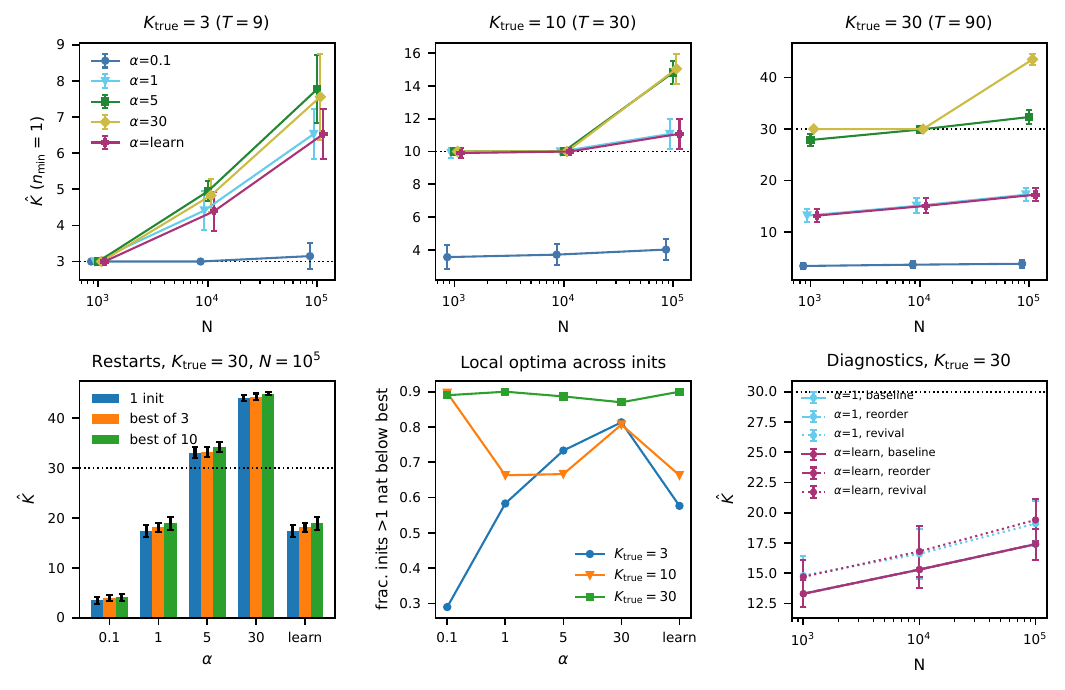}
\caption{$\Khat$ sensitivity. Top: $\Khat$ vs.\ $N$ per $K_{\mathrm{true}}$ (mean $\pm$
sd over 10 data seeds $\times$ 10 initializations; dotted line $K_{\mathrm{true}}$).
Bottom left: restart policies at $K_{\mathrm{true}}{=}30$, $N{=}10^5$. Bottom middle:
fraction of initializations more than 1 nat below their cell's best ELBO. Bottom right:
diagnostic reorder/revival arms at $K_{\mathrm{true}}{=}30$ (experiment-local,
ELBO-guarded; not part of the method).}
\label{fig:khatsens}
\end{figure}

\subsection{$\Khat$ sensitivity: initialization, restarts, concentration, and
order diagnostics (Figure~\ref{fig:khatsens})}
\label{sec:khatsens}
The recovery map of Figure~\ref{fig:f2} fixes one initialization per seed; here we vary
everything it holds fixed. The full factorial crosses
$K_{\mathrm{true}} \in \{3, 10, 30\}$, $N \in \{10^3, 10^4, 10^5\}$,
$\alpha \in \{0.1, 1, 5, 30\}$ plus learned $\alpha$, 10 data seeds, and 10 independent
$k$-means++ initializations per fit (4{,}500 fits), recording $\Khat$ over the full
$n_{\min}$ grid, the entropy-based count, and the converged ELBO. Every fit in this
study reaches the relative-ELBO tolerance: across the $4{,}500$ main fits and the $270$
diagnostic fits below, none of the $4{,}770$ hits the $1{,}000$-cycle limit (worst case
$424$ cycles), so no cell reported here is an artifact of a truncated optimization.
Three summaries matter.
\emph{Initialization:} across the ten $k$-means++ restarts the spread in $\Khat$ is
small (per-cell across-init standard deviation ${\le}{\sim}1.9$ components, and
${\le}{\sim}1.5$ when pooled over the ten seeds and ten initializations of a cell, and $0$
in the well-separated small-$K$, small-$N$ cells), so the recovery
map is not an initialization artifact
even though the ELBO scatter across restarts is large: different initializations reach
different ELBO optima but land on nearly the same $\Khat$. \emph{Restarts:} selecting the
best ELBO of 3 (of 10) restarts does not measurably move $\Khat$ at
$K_{\mathrm{true}}{=}30$, $\alpha{=}1$, $N{=}10^5$ ($17.4 \to 18.1 \to 18.9$ for
best-of-$1/3/10$): the paired change over the 10 data seeds is $+0.7$ components
(95\% CI $[-0.20, 1.60]$) for best-of-3 and $+1.5$ ($[-0.13, 3.13]$) for best-of-10, both
intervals containing zero, even as the selected ELBO improves by $2.6\times10^{4}$ and
$5.3\times10^{4}$ nats. At $\alpha \in \{5, 30\}$ the restart ELBO gains are two to three
orders of magnitude smaller than at $\alpha{=}1$ ($40$--$210$ nats at $N{=}10^5$, against
$2.6\times10^{4}$--$5.3\times10^{4}$) and the best-of-10 change in $\Khat$, while resolved from zero, is
at most about one component ($+1.1$, $[0.47, 1.73]$ at $\alpha{=}5$; $+0.9$,
$[0.37, 1.43]$ at $\alpha{=}30$). Escaping local optima therefore does not close the
$12.6$-component shortfall at $\alpha{=}1$. All intervals in this subsection
are paired $95\%$ $t$ intervals over the 10 data seeds, the arm applied to a common set of
fits.
\emph{Local optima:} flagging an initialization as locally
trapped when its ELBO sits more than 1 nat below the best of its cell,
multi-modality is the rule rather than the exception: at $K_{\mathrm{true}}{=}30$ the
trapped fraction runs $0.81$--$0.90$ across every $\alpha$ and $N$ (at $N{=}10^5$, $0.89$
for $\alpha{=}0.1$ and $0.90$ for $\alpha \in \{1, 5, 30, \text{learn}\}$), and at
$K_{\mathrm{true}}{=}10$ it peaks at $0.90$ for $\alpha{=}0.1$; yet, per the restart result,
the best-ELBO mode carries essentially the same $\Khat$ as the trapped initializations.
Two \emph{diagnostic} arms then probe the mechanism at
$K_{\mathrm{true}}{=}30$, where under-recovery concentrates: every fifth cycle,
(i) \emph{reorder} relabels components by decreasing soft count and recomputes the
stick-breaking posterior for the new order (order-independent ELBO terms are asserted
invariant; the move is kept only if the full ELBO does not decrease;
\citealp{kurihara2007collapsed}), and (ii) \emph{revival} reseeds dead components
($\Nk < 1$) at the worst-explained points of a dominant component and keeps the branch
with the higher ELBO after a short rollout, a guarded split-type move. Both arms live in
experiment code, not in the method; they characterize the failure mode rather than patch
it. Result: reordering is inert (applied about twice per run with zero rejections and
a total ELBO gain of ${\sim}21$ nats; at $N{=}10^5$ the paired change in $\Khat$ from the
baseline $17.4$ is exactly $0$ on all 10 seeds, and the entropy-based count changes by
$-0.003$, $95\%$ CI $[-0.005, -0.001]$: resolved from zero but numerically negligible
against a count of $14.2$), so stick-\emph{order} bias is not the driver;
revival does resurrect a few genuine components ($\Khat: 17.4 \to 19.1$ at $\alpha{=}1$,
paired $+1.7$, $[0.31, 3.09]$; $\to 19.4$ under learned $\alpha$, $+2.0$, $[0.65, 3.35]$;
entropy-$K$ $14.2 \to 16.0$, $+1.77$, $[0.30, 3.24]$) with large ELBO gains
(${\sim}1.1\times10^{4}$ nats, about $3$ applications against $9$ rejections), but still
lands near $19 \ll 30$, and at $\alpha{=}0.1$ it does nothing ($\Khat$ unchanged on every
seed, ELBO gain ${\sim}0$);
neither move closes the $K_{\mathrm{true}}{=}30$ under-recovery gap at $\alpha{=}1$ or
learned $\alpha$. Conclusions that survive all arms: regime dependence of $\Khat$ on
$\alpha$ and its direction; component death as the driver of under-recovery.
What is \emph{not} robust is any specific $\Khat$ value: it drifts with $N$ at fixed
$\alpha$ (the DP is not sample-size invariant, e.g.\ $\alpha{=}30$ recovers
$K_{\mathrm{true}}{=}30$ exactly at $N \le 10^4$ but over-splits to $43.5$ at $N{=}10^5$),
and it moves by about $2$ components under revival ($+2.0$, $95\%$ CI $[0.65, 3.35]$ under
learned $\alpha$); the additional shift under best-of-$10$ restart selection is a point
estimate of $+1.5$ whose interval $[-0.13, 3.13]$ contains zero, so it is a drift we can
neither confirm nor rule out at 10 seeds. We therefore report counts as regime- and
$N$-specific rather than as estimates of a fixed $K$.

\begin{figure}[t]
\centering
\includegraphics[width=0.85\linewidth]{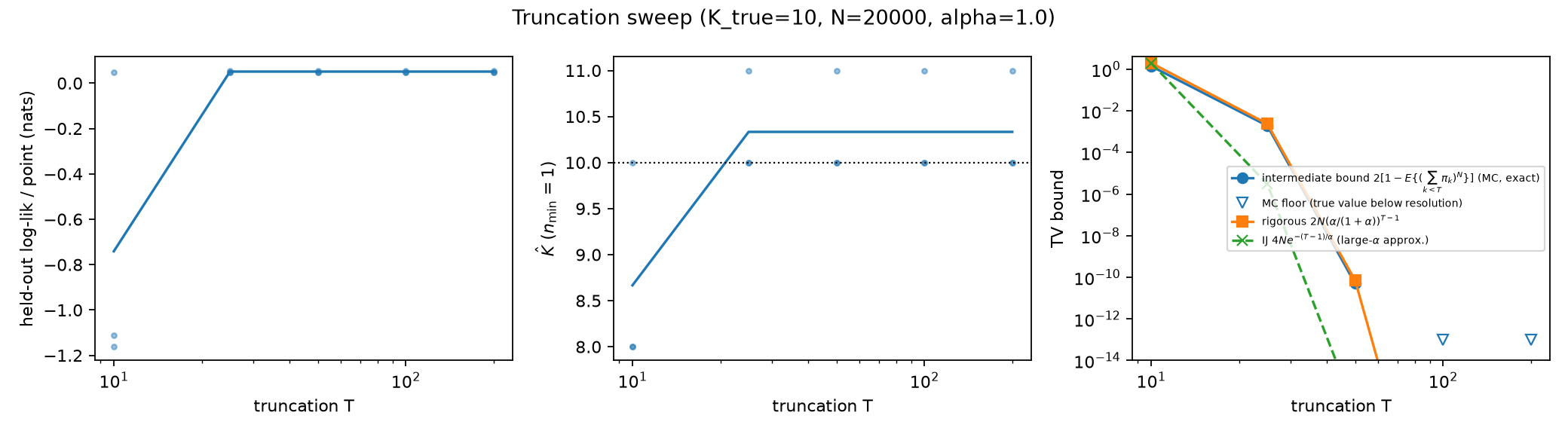}
\caption{Truncation sweep. Left/middle: held-out log-likelihood and $\Khat$ vs.\ $T$
(individual seeds shown; $T{=}K_{\mathrm{true}}$ exhibits bimodal local optima). Right:
the exactly evaluated intermediate bound (MC; censored below MC resolution), the closed-form bound of
Proposition~\ref{prop:truncation}, and the anti-conservative large-$\alpha$ approximation.}
\label{fig:f3}
\end{figure}

\subsection{Truncation is a non-issue, with the right bound (Figure~\ref{fig:f3})}
Sweeping $T \in \{10, 25, 50, 100, 200\}$ at $K_{\mathrm{true}}{=}10$, $N{=}2{\times}10^4$,
$\alpha{=}1$: held-out log-likelihood and $\Khat$ are flat for $T \ge 25$
($\Khat = 10.3 \pm 0.5$); at $T = K_{\mathrm{true}}$ two of three seeds land in degraded
local optima, an optimization effect rather than truncation error. The right panel compares the
exactly evaluated intermediate bound $2(1-\E[(\sum_{k<T}\pi_k)^N])$ with the two closed-form expressions of Proposition~\ref{prop:truncation}: the
rigorous form $2N(\alpha/(1{+}\alpha))^{T-1}$ dominates the exact curve as required, while
the popular $4Ne^{-(T-1)/\alpha}$ approximation lies \emph{below} that exactly evaluated quantity by
3--6 orders of magnitude at $\alpha{=}1$: it undershoots the exact Ishwaran--James
bound it approximates, and so cannot be relied on as a surrogate for that bound, exactly
where these models live.

\begin{figure}[t]
\centering
\includegraphics[width=0.6\linewidth]{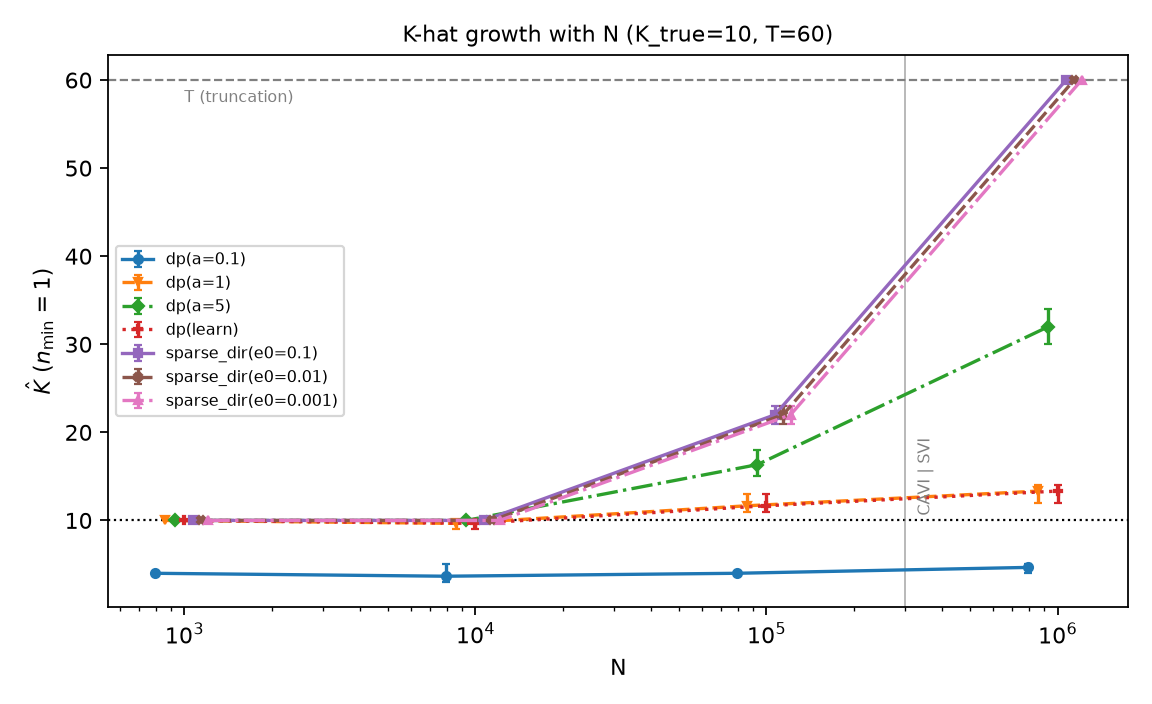}
\caption{$\Khat$ growth with $N$ (3D, $K_{\mathrm{true}}{=}10$, $T{=}60$; CAVI for
$N \le 10^5$, SVI beyond; bars: seed min/max). The sparse finite mixture saturates $T$ by
$N{=}10^6$ for all three $e_0$; the DP variants grow slowly. This inverts the naive reading
of the asymptotic theory (see text).}
\label{fig:f4}
\end{figure}

\subsection{The variational inversion of the $\Khat$ asymptotics (Figure~\ref{fig:f4})}
Asymptotic theory predicts DP posteriors \emph{inflate} the number of components
\citep{miller2014inconsistency} while sparse overfitted mixtures \emph{empty} them
\citep{rousseau2011asymptotic}. Under truncated CAVI on 3D colored mixtures
($K_{\mathrm{true}}{=}10$, $T{=}60$, $N$ from $10^3$ to $10^6$) we observe the opposite
ordering: $\mathrm{dp}(\alpha{=}1)$ grows slowly ($10 \to 13.3$ over three decades of $N$,
consistent with log-like inflation), while the sparse variant saturates the \emph{entire}
truncation ($\Khat = 22$ at $10^5$, $60 = T$ at $10^6$), identically for
$e_0 \in \{0.1, 0.01, 0.001\}$. The $e_0$-insensitivity is diagnostic: the posteriors do
differ by exactly the $e_0$ perturbation while $\Khat$ does not, so the over-splitting is
driven by likelihood-dominated coordinate-ascent dynamics (splitting a dense cluster costs
almost nothing at large $N$, and CAVI has no merge moves), not by the prior. Conversely, the
DP's size-biased stick ordering suppresses late components early, which resists
over-splitting at large $N$ but also under-recovers when $\alpha$ is too small for the
regime ($\mathrm{dp}(\alpha{=}0.1)$ merges to $\Khat \approx 4$ here). At these sample
sizes, \emph{fitting dynamics dominate prior asymptotics}.

\begin{figure}[t]
\centering
\includegraphics[width=0.46\linewidth]{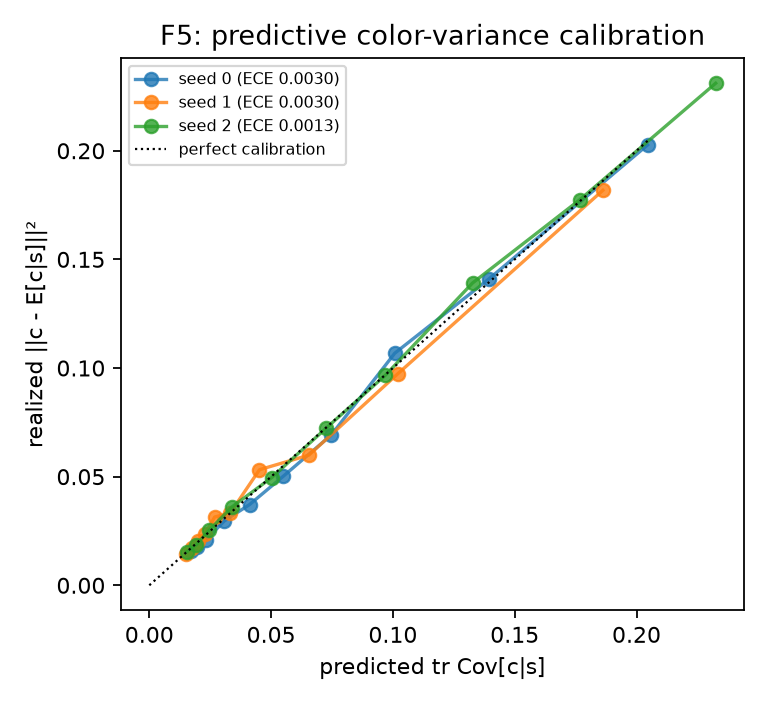}\hfill
\includegraphics[width=0.42\linewidth]{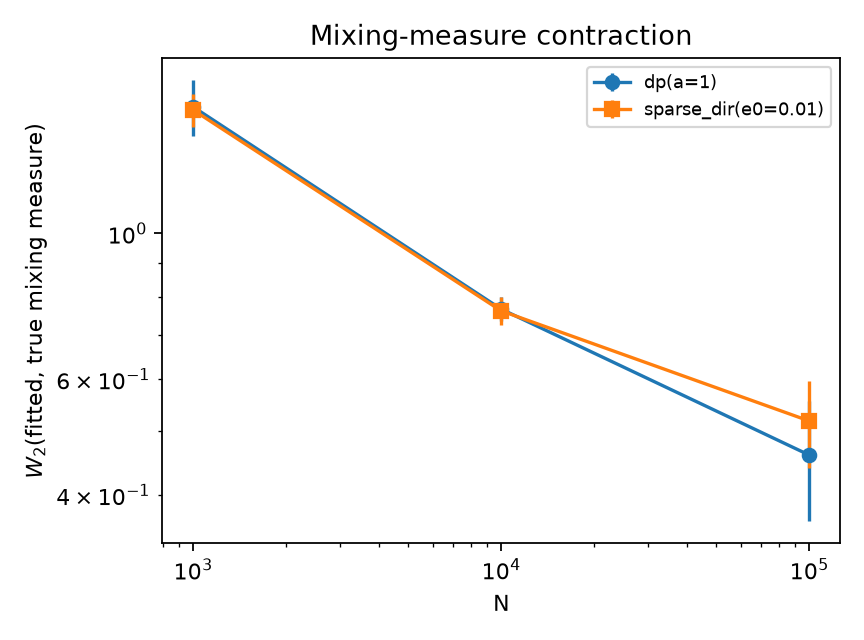}
\caption{Left: predictive color-variance calibration, binned realized squared error vs.\
binned predicted variance (three seeds); ``ECE'' denotes the variance-calibration analogue
$\sum_b f_b\,\lvert \mathrm{MSE}_b - \overline{\mathrm{Var}}_b\rvert$
\citep{kuleshov2018accurate}. Right: $W_2$(fitted, true mixing measure) vs.\ $N$ (three
seeds; atoms in joint spatial--color mean space); contraction theory for exact posteriors
is given by \citet{nguyen2013convergence}, and no claim is made that CAVI attains those
rates.}
\label{fig:f5}
\end{figure}

\subsection{Calibrated color uncertainty (Figure~\ref{fig:f5})}
On 3D synthetics with overlapping clusters ($3\sigma$ separation), so that predictive
color uncertainty genuinely varies in space through component ambiguity, we bin held-out points by predicted
$\operatorname{tr}\mathrm{Cov}[c\,|\,s^\ast]$ (10 quantile bins) and compare each bin's mean
predicted variance to its realized $\lVert c^\ast - \E[c|s^\ast]\rVert^2$. The calibration
curve sits on the diagonal across the full variance range; the regression expected
calibration error is $1.3$--$3.0 \times 10^{-3}$ (three seeds), with global mean predicted
variance within $2\%$ of realized MSE. This is the well-specified regime, where good
calibration is expected; under misspecification (the real scenes of
\S\ref{sec:scenes}, same protocol on held-out frames) calibration degrades to ECE
$0.011$ (\emph{lego}, variance over-estimated by $18\%$) and $0.016$ (\emph{chair},
under-estimated by $21\%$): still the right order of magnitude and positively correlated
with realized error, but no longer sharp. \S\ref{sec:uncertainty8} extends this
misspecified assessment to all eight scenes with coverage and predictive likelihood, and
Appendix~\ref{app:calcurves} plots the corresponding eight-scene calibration curves
in the same binning.
In fixed-color-precision baselines the corresponding quantity is constant and
non-adaptive, so it cannot be calibrated against realized error in either regime.

\subsection{Mixing-measure contraction (Figure~\ref{fig:f5}, right)}
$W_2$ between the fitted mixing measure (posterior mean atoms, pruned expected weights) and
the truth contracts as a power law in $N$ ($1.56 \to 0.46$ for $\mathrm{dp}$, $\alpha{=}1$,
over $N = 10^3 \to 10^5$); at $N = 10^5$ the DP variant is closer than the sparse variant
($0.46$ vs.\ $0.52$ mean), consistent with Figure~\ref{fig:f4}: over-split components carry
weight off the true atoms.

\subsection{Images at matched effective budgets: a deconfounded comparison
(Table~\ref{tab:images})}
Following VBGS's image protocol (Tiny-ImageNet per \citealp{le2015tiny}; $64{\times}64$,
pixels as points, their data pipeline and pure-JAX renderer for \emph{all} methods), we fit
DP-Splat with $T{=}2000$ (VBGS's demo budget) on 12 validation images and compare against
three baselines: (a) VBGS at its full budget $K{=}2000$; (b) VBGS at fixed $K = \Khat$
(matched effective budget, VBGS's own single-pass protocol); and (c) a \emph{converged}
fixed-$K$ Dirichlet CAVI fit at $K = \Khat$; baseline (c) isolates the weight prior from
the effect of running coordinate ascent to convergence. The result is clean: DP-Splat
exceeds single-pass VBGS at matched $\Khat$ by $+2.7$\,dB on average, but against the
equally converged fixed-$K$ baseline the difference essentially vanishes
($-0.01$ and $+0.17$\,dB mean at $\alpha \in \{1, 100\}$). \emph{The DP prior's
contribution is selecting $\Khat$, not per-component rendering efficiency at a given
$K$}; the raw
matched-budget gap over VBGS is attributable to convergence. The predictive
uncertainty machinery is not the DP prior's contribution either: it is the conjugate NIW
color model, which the fixed-$K$ Dirichlet baseline shares and which calibrates equally
well there (\S\ref{sec:uncertainty8}). Two further
observations: (i) $\Khat$ stays two orders of magnitude below the $T{=}2000$ budget
regardless of $\alpha$ ($\Khat \in [9, 17]$ at $\alpha{=}1$, $[29, 53]$ at $\alpha{=}100$;
component death again, the prior alone cannot force capacity into a fit), so absolute PSNR
sits $3$--$10$\,dB below VBGS's full-2000-budget render; (ii) at $\alpha = 1$ the DP
prior's own $\E[K] \approx \log N \approx 8$ is faithfully realized: the model does what
its prior says.

\begin{table}[t]
\centering
\caption{Images at matched effective budgets (12 Tiny-ImageNet images, $T{=}2000$;
mean [min, max] over images; all methods rendered identically; ``conv.\ dir@$\Khat$'' =
converged fixed-$K$ Dirichlet CAVI at the same $\Khat$, the deconfound).}
\label{tab:images}
\begin{tabular}{lcccc}
\toprule
$\alpha$ & $\Khat$ & vs.\ VBGS@$\Khat$ (single-pass) & vs.\ conv.\ dir@$\Khat$ &
vs.\ VBGS@2000 \\
\midrule
1 & 9--17 & $+2.80$ $[+0.64, +4.77]$ & $-0.01$ $[-0.77, +0.77]$ & $-6.47$ \\
100 & 29--53 & $+2.70$ $[+0.51, +5.46]$ & $+0.17$ $[+0.02, +0.70]$ & $-4.79$ \\
\bottomrule
\end{tabular}
\end{table}

\subsection{3D scenes, continual fitting, and scaling (Figure~\ref{fig:phase3})}
\label{sec:scenes}
On NeRF-synthetic scenes \citep[test-split RGB-D point clouds; 40 frames $\times$ 20k
subsampled points; matched budget $T = K = 2000$]{mildenhall2020nerf}, DP-Splat is fit by SVI over the frame stream (10
epochs) and VBGS by its own continual protocol. We report the rasterizer-free held-out
metric available to both methods, point-level color prediction $\E[c\,|\,s]$ against
ground truth on a single unseen frame per scene, plus $\Khat$ trajectories and DP-Splat's predictive
variance maps (Table~\ref{tab:scenes}). At $\alpha{=}100$, fixed in advance, DP-Splat's
held-out prediction stays within $0.33$\,dB of VBGS's on all eight scenes and is ahead on
four of them (\emph{hotdog} $+1.31$, \emph{ship} $+0.31$, \emph{lego} $+0.29$,
\emph{materials} $+0.03$\,dB; \S\ref{sec:uncertainty8}), while using $3.9$--$7.9\times$
fewer occupied components, mean $5.0\times$. Those are single-run margins: over three
seeds the first three of the four are resolved at two standard errors and
\emph{materials} is a tie (\S\ref{sec:uncertainty8}). Both counts apply the same occupancy rule,
$\Nk > 1$ on a full-data E-step. This matters for the ratio: the VBGS reference
implementation instead counts every component with $\Nk > 0$, which credits the baseline
with ${\sim}1373$ components rather than the ${\sim}1291$ it actually occupies under the
rule used for $\Khat$, and inflates the ratio by about $6\%$.
Compute is not the driver: a \emph{single-pass} DP-Splat fit
(23--25\,s, comparable to VBGS's 36\,s one-shot protocol) already reaches 17.25\,dB on
\emph{lego} and 16.03\,dB on \emph{chair}, within $0.17$ and $0.01$\,dB of the
ten-epoch fits at about six times the wall-clock (Table~\ref{tab:scenes}). The
fitted complexity also orders the two scenes plausibly: the geometrically richer
\emph{lego} receives more components than \emph{chair} at identical settings (an $n{=}2$
observation we report as such; \S\ref{sec:uncertainty8} extends the scene set), and, unlike
the baseline, whose color variance is constant and non-adaptive, every prediction carries an
inferred predictive variance (Figure~\ref{fig:phase3}, Appendix~\ref{app:chair}: uncertainty
concentrates on edges and occlusion boundaries). Unlike on $64{\times}64$ images, $\Khat$ here rises far past 30 with
$\alpha$; the earlier saturation was data-complexity-limited, not a universal ceiling.
Rasterized novel-view PSNR/SSIM require the CUDA rasterizer and full-resolution protocols
(\S\ref{sec:raster}); published full-pipeline
3DGS numbers (e.g., \emph{chair} $35.8$\,dB; \citealp{kerbl2023gaussian}) are not comparable
to this point-level proxy. Finally, the natural-gradient variant's per-step cost is flat in
$N$ (0.04--0.05\,s/step at $|B| = 2^{16}$, $T{=}60$, from $N = 10^5$ to $10^7$ on CPU):
dataset size is not the binding constraint for DP-Splat.

\begin{table}[t]
\centering
\caption{Scenes at matched budget $T = K = 2000$ (a single held-out frame per scene,
under the reference implementation's protocol; point-level metrics;
point-PSNR in original $[0,1]$ color units; wall-clock on the same CPU; $\alpha{=}100$
fixed a priori, no per-scene tuning). The likelihood column is the \emph{joint}
per-point held-out density $\log \hat p(s, c)$ over position and color, in normalized
units. Table~\ref{tab:uncertainty} reports the \emph{conditional}
$\log \hat p(c \mid s)$ for the same fits, so the two tables give different values for
the same $\alpha{=}100$ arms. VBGS's color precision is fixed, so its predictive
variance and likelihood are constant and non-adaptive rather than inferred; we leave those
entries blank rather than report a quantity that carries no fitted information.
DP-Splat's $\Khat$ is computed from a full-data E-step at the final posterior. The
two VBGS entries are its occupancy under its reference implementation's rule, $N_k > 0$,
and under the $N_k > 1$ rule that defines $\Khat$ throughout this paper; component-count
ratios between the two models are only meaningful under the second. The one-epoch rows
report the running single-minibatch estimate and are not directly comparable to the
ten-epoch rows.}
\label{tab:scenes}
\begin{tabular}{llcccc}
\toprule
Scene & Model & Effective $K$ & Point-PSNR (dB) & Joint LL/pt & Fit (s) \\
\midrule
lego & DP-Splat $\alpha{=}1$ (10 ep.) & 20 & 14.56 & $-1.65$ & 159 \\
lego & DP-Splat $\alpha{=}100$ (1 ep.) & 276 & 17.25 & $+0.69$ & 25 \\
lego & DP-Splat $\alpha{=}100$ (10 ep.) & \textbf{260} & \textbf{17.42} & $\mathbf{+1.03}$ & 155 \\
lego & VBGS ($K{=}2000$) & 1374 / 1292 & 17.13 & n/a & 36 \\
\midrule
chair & DP-Splat $\alpha{=}1$ (10 ep.) & 19 & 14.55 & $-4.68$ & 152 \\
chair & DP-Splat $\alpha{=}100$ (1 ep.) & 264 & 16.03 & $-2.06$ & 23 \\
chair & DP-Splat $\alpha{=}100$ (10 ep.) & 230 & 16.03 & $-1.95$ & 153 \\
chair & VBGS ($K{=}2000$) & 1373 / 1291 & 16.17 & n/a & 35 \\
\bottomrule
\end{tabular}
\end{table}

\subsection{Uncertainty on all eight scenes: calibration, coverage, predictive
likelihood (Table~\ref{tab:uncertainty})}
\label{sec:uncertainty8}
We extend the misspecified-regime uncertainty assessment from two scenes to all eight
NeRF-synthetic scenes, with the training protocol of \S\ref{sec:scenes} unchanged and the
full $\alpha$ grid $\{1, 10, 100, 1000\}$ per scene. For each fit, on the held-out frame:
the variance-decile regression-ECE of Figure~\ref{fig:f5}; empirical central-interval
coverage of the per-dimension moment-matched Gaussian intervals from
$(\E[c|s], \mathrm{Var}[c|s])$ at the $68.3/90/95\%$ levels, with a coverage-ECE
(mean absolute nominal-vs-empirical gap over a 19-level grid); and the exact held-out
conditional log-likelihood $\log \hat p(c \mid s)$ under the spatially reweighted
mixture-of-Students predictive. Table~\ref{tab:uncertainty} reports the a-priori-fixed
$\alpha{=}100$ setting per scene; Appendix~\ref{app:uncertainty} tabulates the full grid
and plots the coverage curves, and Appendix~\ref{app:calcurves} the binned
variance-calibration curves behind the ECE column. Findings: calibration is good in aggregate but never
perfect under misspecification. Regression-ECE is small on every scene and setting
(${\le}0.031$, with the maximum $0.0303$ on \emph{drums} at $\alpha{=}1$); six of the eight
scenes minimize it at $\alpha \in [100, 1000]$, but \emph{ficus} and \emph{lego} minimize it
at $\alpha{=}1$, so a large concentration is not uniformly the best-calibrated setting.
Conditional log-likelihood rises monotonically with $\alpha$ on
every scene and held-out point-PSNR does so on seven of eight, both saturating by
$\alpha{=}1000$. Those are single-run orderings, and the one exception, a $0.12$\,dB fall
on \emph{lego} between $\alpha{=}100$ and $\alpha{=}1000$, is a seed effect rather than a
property of the model: over three seeds \emph{lego} rises from $17.29 \pm 0.12$ to
$17.47 \pm 0.17$\,dB, and the $\alpha{=}100 \to 1000$ increment in point-PSNR is positive
on all eight scenes, resolved at two standard errors on four of them (\emph{drums},
\emph{hotdog}, \emph{mic}, \emph{ship}). Monotonicity in $\alpha$ should be read in
seed-mean terms.
The parity claim against VBGS needs the same treatment; its evidence basis is, like
every number in this subsection, a single held-out frame per scene under the reference
implementation's protocol, in explicit contrast to the 25-view rasterized evaluation of
\S\ref{sec:raster}. Repeating the $\alpha{=}100$ and
$\alpha{=}1000$ arms and the VBGS baseline three times per scene, with every random element
reseeded, including the $k$-means initialization that earlier runs held fixed, gives
per-scene per-arm point-PSNR standard deviations of $0.003$--$0.166$\,dB; the point-PSNR
entries of Table~\ref{tab:uncertainty} are the first of these three seeds. On those seed
means, at $\alpha{=}1000$ DP-Splat exceeds VBGS by more than two standard errors of the
difference on four scenes (\emph{hotdog} $+1.58$, \emph{lego} $+0.35$, \emph{ship}
$+0.28$, \emph{drums} $+0.09$\,dB), falls below it on one (\emph{ficus} $-0.16$\,dB), and
is statistically tied on the remaining three (\emph{chair} $-0.14$, \emph{materials}
$+0.06$, \emph{mic} $-0.05$\,dB, each inside its two-sigma band).
Central-interval coverage is close to nominal at the
$90/95\%$ levels on most scenes, but at $68.3\%$ the intervals are too narrow on
\emph{chair}, \emph{ficus} and \emph{mic}, which \emph{under-cover} at every $\alpha$
($0.59$--$0.64$ at $\alpha{=}100$ against the nominal $0.683$): the model is
\emph{over}-confident on those three scenes.
Sharpness and interval calibration are not traded off in a single direction. Raising
$\alpha$ from $1$ to $1000$ \emph{improves} coverage-ECE on four scenes (\emph{ship}
$0.071 \to 0.032$, \emph{materials} $0.034 \to 0.016$, \emph{mic} $0.054 \to 0.040$,
\emph{chair} $0.070 \to 0.059$, the last of these monotonically) and \emph{worsens} it on
the other four, monotonically in each (\emph{ficus} $0.012 \to 0.056$, \emph{hotdog}
$0.059 \to 0.101$, \emph{drums} $0.016 \to 0.052$, \emph{lego} $0.034 \to 0.062$). The
$\alpha$ that optimizes sharpness is therefore not the one that optimizes interval
calibration on half the scene set, and the sign of the effect is scene-dependent.

\paragraph{Matched-capacity ablation: which prior supplies the calibration?}
The predictive variance above is produced by the conjugate NIW color factor, which the
weight prior does not touch, so we tested directly whether the DP prior contributes any of
it. For each scene we fit a fixed-$K$ symmetric Dirichlet mixture ($e_0 = 1/K$) at exactly
the capacity the DP selects at $\alpha{=}100$, that is $K = \Khat$ from
Table~\ref{tab:uncertainty}, under an SVI protocol identical in every other respect, and
paired it seed by seed against the DP arm over three seeds (8 scenes $\times$ 3 seeds per
arm; per-scene results in Table~\ref{tab:dirablation}). The outcome is a null on every
uncertainty metric. Averaged over scenes, the fixed-$K$ arm differs from the DP arm by
$-5\times10^{-5}$ in regression-ECE (paired $t$ over the eight scenes, $p = 0.86$, 95\%
interval $[-7.2\times10^{-4}, +6.1\times10^{-4}]$; $5\%$ of
the mean per-scene seed spread of $1.1\times10^{-3}$), by $-9\times10^{-5}$ in coverage-ECE
($p = 0.89$; $[-1.6\times10^{-3}, +1.4\times10^{-3}]$; $2\%$ of a spread of
$4.4\times10^{-3}$), and by $-0.010$ nats in conditional
held-out log-likelihood ($p = 0.44$; $[-0.038, +0.019]$). Point-PSNR is the one metric on which the two arms
are not interchangeable: the DP arm is ahead by $0.077$\,dB ($p = 0.022$). We do not
present that as resolved. It fails a Bonferroni bar of $0.0125$ for the four tests reported
here and it sits below the mean per-scene seed spread of $0.121$\,dB, so we read it as a
small advantage in a consistent direction on most scenes rather than a settled
difference at three seeds. The aggregate null is a null with scene-level texture rather
than a suppressed consistent effect: two scenes separate by more than twice their own seed
spread and both favor the DP arm (\emph{ficus} regression-ECE, $+1.26\times10^{-3}$ or
$2.9$ times its spread, and \emph{drums} point-PSNR at $2.1$ times its spread), and the
heterogeneity cancels in the mean. The realized capacities are worth stating, because the
asymmetry runs against the DP arm's PSNR edge rather than for it: scored by the same
$\Nk > 1$ rule, the fixed-$K$ arm occupies $86.7$--$98.3\%$ of the budget it was given
(mean $93.6\%$), and because $K$ is pinned to the seed-0 DP count while the DP arm's own
occupancy drifts upward with seed, the fixed-$K$ arm ends with fewer live components than
the DP arm on seven of the eight scenes ($83$--$100\%$ of the DP count, mean $93\%$). The
calibration null therefore holds at slightly lower realized capacity, and the $0.077$\,dB
PSNR difference is if anything partly a capacity difference rather than a prior effect. As
single-seed supporting evidence only, sweeping the
fixed-$K$ arm over $K \in \{\Khat/4, \Khat/2, \Khat, 2\Khat\}$ shows that at and above the
operating capacity its regression-ECE agrees with the DP arm's ECE-versus-capacity curve,
interpolated to the same occupied-component count, to a mean absolute $9\times10^{-4}$ and
a worst case of $3.4\times10^{-3}$, the mean being about the seed noise of the metric;
below the operating point the capacity sensitivity is
strongly scene-dependent (ECE range over the four capacities from $1.0\times10^{-3}$ on
\emph{ship} to $1.1\times10^{-2}$ on \emph{drums}, with a quarter-capacity outlier on
\emph{lego} at ECE $0.0213$ of the local-optimum kind documented in \S\ref{sec:khatsens}).
That sweep is one seed per cell and we build nothing on it alone. The conclusion we draw
is the precise form of contribution~4: calibrated predictive color uncertainty is a
property of the conjugate NIW color model, which a fixed-$K$ Dirichlet mixture at matched
capacity reproduces at three seeds with no resolved difference (the paired 95\% intervals
above all cover zero), and the DP prior's distinct
contribution is automatic complexity
selection, the step that supplies the $K$ the fixed-$K$ arm has to be told.

\begin{table}[t]
\centering
\caption{Expanded uncertainty evaluation, all eight NeRF-synthetic scenes
($\alpha{=}100$ fixed a priori; a single held-out frame per scene, under the reference
implementation's protocol; point space; PSNR and regression-ECE in
original $[0,1]$ color units; coverage per color dimension at nominal $68.3/90/95\%$ and
scale-free; conditional held-out log-likelihood $\log \hat p(c \mid s)$ per point in
normalized units, as in Table~\ref{tab:uncertaintyfull}, and not the joint quantity of
Table~\ref{tab:scenes}; VBGS point-PSNR at matched budget
for reference). Full $\alpha$ grid: Appendix~\ref{app:uncertainty}.}
\label{tab:uncertainty}
\small
\begin{tabular}{lcccccc}
\toprule
Scene & $\Khat$ & PSNR & ECE & Cov.\ 68/90/95 (\%) & Cond.\ LL/pt & VBGS PSNR \\
\midrule
chair     & 230 & 16.03 & 0.0158 & 58.7/84.8/92.2 & $-0.49$ & 16.17 \\
drums     & 300 & 16.29 & 0.0124 & 75.6/90.7/93.5 & $+2.33$ & 16.39 \\
ficus     & 322 & 20.04 & 0.0075 & 61.6/85.9/91.4 & $-0.28$ & 20.37 \\
hotdog    & 261 & 24.13 & 0.0061 & 81.3/94.1/96.1 & $+2.04$ & 22.83 \\
lego      & 260 & 17.42 & 0.0109 & 78.2/93.1/96.2 & $+2.78$ & 17.13 \\
materials & 315 & 16.00 & 0.0048 & 70.7/91.1/95.0 & $+1.57$ & 15.97 \\
mic       & 164 & 17.88 & 0.0128 & 63.6/88.0/92.6 & $+1.46$ & 18.12 \\
ship      & 330 & 20.04 & 0.0046 & 74.7/92.1/95.0 & $-0.15$ & 19.72 \\
\bottomrule
\end{tabular}
\end{table}

\begin{figure}[t]
\centering
\includegraphics[width=\linewidth]{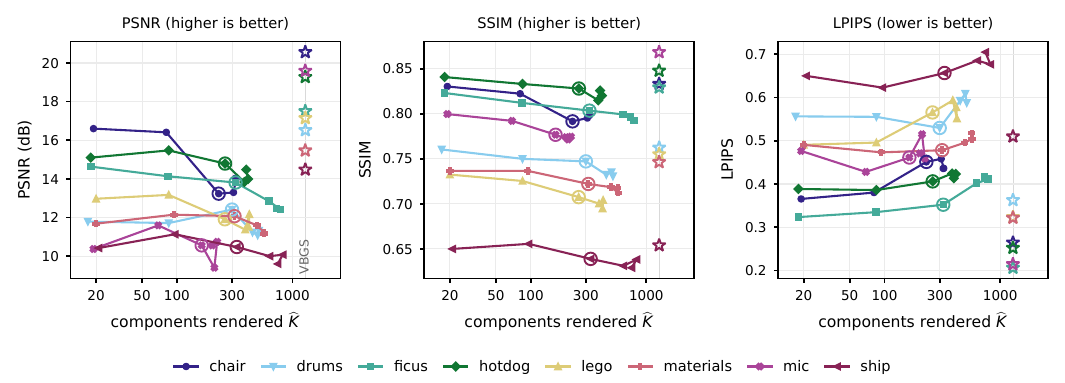}
\caption{Rasterized quality versus the number of components actually rendered, $\Khat$
(\texttt{gsplat}; $25$ strictly held-out views per scene; log $x$-axis). One line per scene,
its markers ordered by increasing $\alpha \in \{1, 10, 100, 10^3, 10^4, 10^5\}$, which runs
left to right except at $\alpha{=}10^5$, where seven of the eight scenes step back to the
left as the posterior sheds components; the ringed
marker is the a-priori $\alpha{=}100$ arm tabulated in Table~\ref{tab:raster}, and the open
star is the same scene's fixed-$K$ VBGS operating point inside the grey band at
${\sim}1290$ components. Within each scene rendered quality is flat to declining in
$\Khat$: PSNR correlates negatively with $\Khat$ on all eight scenes ($r$ from $-0.39$ to
$-0.99$, mean $-0.76$) and so does SSIM ($r$ from $-0.85$ to $-0.98$, mean $-0.94$), while
LPIPS, for which lower is better, correlates positively on all eight ($r$ from $+0.37$ to
$+0.99$). The best-rendering arm is never the largest-$\alpha$ arm. The VBGS budget sits far
to the right of every DP arm, but over the range swept, adding components does not move
DP-Splat toward it.}
\label{fig:r3quality}
\end{figure}

\subsection{Rasterized novel-view evaluation (Table~\ref{tab:raster},
Figure~\ref{fig:r3quality})}
\label{sec:raster}
The evaluations above are point-space by design; this subsection reports the rendered
counterpart on all eight NeRF-synthetic scenes. The fitted posteriors are mapped to splats
exactly as in \S\ref{sec:inference} (Cholesky scale/rotation decomposition of
$\E[\Sigma_k]$, binary opacity on the live components, scale modifier $1.41$) and rendered
with the \texttt{gsplat} rasterizer against a white background, mirroring the VBGS render
protocol.

\textbf{Evaluation views are strictly held out.} Each fit consumes the first $40$
frames of a scene's $200$-frame split, and the $25$ scored views are sampled uniformly from
frame $40$ onward, so no scored view was seen during fitting. We state this in the body
rather than a footnote because the alternative, sampling the $25$ views uniformly across
the whole $200$-frame split while training on frames $0$--$39$, would place $5$ of the $25$
scored views inside the training set. The live-component mask is likewise read from a
full-data E-step rather than from the final stochastic minibatch, whose count can fall as
much as $22\%$ below the full-data value (\emph{chair} at $\alpha{=}10^3$: $241$ against
$309$). Both choices are the conservative ones for DP-Splat: each moves the comparison
against it relative to the alternative.

We sweep $\alpha \in \{1, 10, 100, 10^3, 10^4, 10^5\}$. The primary comparison in
Table~\ref{tab:raster} is the $\alpha{=}100$ arm, the same setting fixed in advance for the
point-space evaluation of Table~\ref{tab:uncertainty}, against the fixed-$K$ VBGS baseline
at its ${\sim}1290$-component budget. Rendered PSNR is non-monotone in $\alpha$ and never
peaks at the largest $\alpha$ on any scene.

The finding is unambiguous:
at the a-priori $\alpha{=}100$ setting VBGS wins every scene on all three metrics.
DP-Splat renders $3.40$ (\emph{materials}) to $9.03$\,dB (\emph{mic}) below VBGS, mean
$5.16$\,dB, while compositing $3.9$ to $7.9\times$ fewer components (mean $5.0\times$).
Across the whole six-arm sweep the ordering has a single exception, \emph{ship} at
$\alpha{=}10$, whose SSIM of $0.656$ edges VBGS's $0.654$; PSNR and LPIPS favor VBGS at
every arm on every scene. Selecting the best arm per scene from the sweep would
narrow the mean deficit to $4.17$\,dB (range $2.86$ to $8.00$\,dB), but that selection is
made on the test set, so it is an oracle upper bound on DP-Splat rather than a configuration
anyone could deploy; we quote it once, in the caption of Table~\ref{tab:raster}, and rest no
claim on it.

The sweep does not license attributing this gap to the pruning that makes DP-Splat compact.
The data-supported $\Khat$ does saturate in $\alpha$ well below the truncation and far below
the VBGS budget (seven of eight scenes reach their maximum live-component count at
$\alpha{=}10^4$ and then shed components at $\alpha{=}10^5$; \emph{drums} alone is still
adding components at $\alpha{=}10^5$ ($500{\to}514$); \emph{mic} never exceeds $223$
components at any $\alpha$), so the renderer composites between $17$ and $824$ Gaussians
against VBGS's ${\sim}1290$. But the extra components do not buy quality. This is
the same model whose held-out point-space color prediction ---a single held-out frame
per scene, over three seeds--- is at parity with VBGS
(\S\ref{sec:scenes}, \S\ref{sec:uncertainty8}); point-space color prediction and rasterized
sharpness are distinct measurements, and the model is competitive on the first and loses on
the second.

Figure~\ref{fig:r3quality} plots the quality-versus-$\Khat$ frontier, and it runs against
the component-budget reading. Within each scene, across the six-arm $\alpha$ sweep,
rasterized quality is flat to declining in $\Khat$: the Pearson correlation between $\Khat$
and PSNR is negative on all eight scenes ($r$ from $-0.39$ to $-0.99$, mean $-0.76$), the
correlation with SSIM is negative on all eight ($r$ from $-0.85$ to $-0.98$), and the
correlation with LPIPS is positive on all eight ($r$ from $+0.37$ to $+0.99$), which agrees
in sign because lower LPIPS is better. The decline is not monotone arm by arm; several
scenes show SSIM upticks at the largest $\alpha$. The best-rendering arm is never the
largest-$\alpha$ arm on any scene. Over the range of $\Khat$ the $\alpha$ sweep reaches,
then, spending more components does not move DP-Splat toward VBGS: the sweep lends no
support to a component-budget account of the gap. Two candidates the sweep alone cannot separate from
that conclusion are worth naming, both under our own control. First, $\Khat$ is varied only by
varying $\alpha$, so the component count co-varies with the entire posterior rather than
being manipulated on its own. Second, the export map is held fixed across arms and is not
calibrated per component count: binary opacity on the live mask and a scale modifier of
$1.41$, inherited from the VBGS render protocol, are applied identically at $17$ and at
$824$ components, and a fixed opacity-and-scale map that suits one occupancy need not suit
another. Both are
tested directly by the controlled decomposition that closes this subsection: the first
is removed (the negative quality--capacity relation survives manipulating $K$ on its own),
and the second turns from a caveat into a measured multi-dB contributor.

\begin{table}[t]
\centering
\caption{Rasterized novel-view evaluation on the eight NeRF-synthetic scenes
(\texttt{gsplat} renderer; white background; scale modifier $1.41$; binary opacity on live
components; $25$ evaluation views per scene, sampled from frame $40$ onward and therefore
strictly disjoint from the $40$ training frames). DP-Splat is shown at $\alpha{=}100$, the
concentration fixed in advance for Table~\ref{tab:uncertainty} and \emph{not} chosen on the
test set, against the fixed-$K$ VBGS baseline at its ${\sim}1290$-component budget. Higher
PSNR/SSIM and lower LPIPS are better; $\Delta$PSNR is VBGS minus DP-Splat, and the last
column is how many times more components VBGS composites. VBGS wins every scene on every
metric, by $3.40$--$9.03$\,dB while compositing $3.9$--$7.9\times$ as many components. For
reference only: picking the best-PSNR arm per scene from the $\alpha$ sweep would reduce the
mean deficit from $5.16$ to $4.17$\,dB ($2.86$--$8.00$\,dB), but that arm is selected on the
test set and is an oracle upper bound on DP-Splat, not an achievable operating point.}
\label{tab:raster}
\small
\begin{tabular}{lrrrrrrrrrr}
\toprule
& \multicolumn{4}{c}{DP-Splat ($\alpha{=}100$)} & \multicolumn{4}{c}{VBGS (fixed $K$)}
& \multicolumn{2}{c}{Gap} \\
\cmidrule(lr){2-5} \cmidrule(lr){6-9} \cmidrule(lr){10-11}
Scene & $\Khat$ & PSNR & SSIM & LPIPS & $K$ & PSNR & SSIM & LPIPS
& $\Delta$PSNR & $\times$ comp. \\
\midrule
chair     &  230 & 13.24 & 0.792 & 0.451 & 1291 & 20.56 & 0.833 & 0.265 & 7.32 & $5.6\times$ \\
drums     &  300 & 12.42 & 0.747 & 0.530 & 1292 & 16.52 & 0.762 & 0.363 & 4.10 & $4.3\times$ \\
ficus     &  322 & 13.81 & 0.804 & 0.352 & 1291 & 17.50 & 0.829 & 0.206 & 3.69 & $4.0\times$ \\
hotdog    &  261 & 14.80 & 0.828 & 0.406 & 1290 & 19.27 & 0.848 & 0.252 & 4.48 & $4.9\times$ \\
lego      &  260 & 11.91 & 0.707 & 0.565 & 1292 & 17.15 & 0.755 & 0.323 & 5.23 & $5.0\times$ \\
materials &  315 & 12.07 & 0.722 & 0.478 & 1290 & 15.47 & 0.746 & 0.322 & 3.40 & $4.1\times$ \\
mic       &  164 & 10.56 & 0.777 & 0.461 & 1290 & 19.60 & 0.868 & 0.214 & 9.03 & $7.9\times$ \\
ship      &  330 & 10.48 & 0.639 & 0.657 & 1291 & 14.48 & 0.654 & 0.510 & 4.01 & $3.9\times$ \\
\midrule
mean      &  273 & 12.41 & 0.752 & 0.488 & 1291 & 17.57 & 0.787 & 0.307 & 5.16 & $5.0\times$ \\
\bottomrule
\end{tabular}
\end{table}

\textbf{Prior, capacity, and export map: a controlled decomposition.} The two
candidates just named are not left standing; each is varied directly, holding the rest of
the render protocol fixed (same \texttt{gsplat} configuration, white background, the same
$25$ strictly held-out views), with full per-scene numbers in
Table~\ref{tab:rendercontrols} (Appendix~\ref{app:rendercontrols}). The first control
targets the weight prior itself. A fixed-$K$ symmetric Dirichlet mixture ($e_0 = 1/K$) fit
at $K = \Khat$, the capacity DP-Splat selects at $\alpha{=}100$ --- the render-side
counterpart of the point-space ablation of Appendix~\ref{app:dirablation} --- renders
within $+0.35$\,dB of DP-Splat on average over three seeds (per-scene mean differences
$-0.36$ to $+1.68$\,dB), and both arms trail VBGS by nearly the same margin: mean
$4.81$\,dB for the fixed-$K$ arm against $5.16$\,dB for DP-Splat. A model with no
stick-breaking prior, handed the same capacity and the same export map, lands in the same
place, so the deficit is not attributable to the weight prior.

The second control varies capacity directly. Sweeping the same fixed-$K$ arm over
$K \in \{\Khat/4, \Khat/2, \Khat, 2\Khat\}$ (seed $0$; the realized live count falls below
the nominal capacity at the larger arms, e.g.\ \emph{ship} occupies $549$ of $660$)
reproduces the negative quality--capacity relation with $K$ manipulated on its own rather
than through $\alpha$: rendered PSNR peaks at $\Khat/2$ or $\Khat/4$ on all eight scenes,
never at $\Khat$ or $2\Khat$, falls by $0.68$ to $2.41$\,dB from the best arm to the
$2\Khat$ arm, and correlates negatively with the realized component count on all eight
scenes ($r$ from $-0.48$ to $-0.96$, mean $-0.82$). This removes the first confound
disclosed above: the flat-to-declining pattern of Figure~\ref{fig:r3quality} is a property
of component count under this fitting-and-export pipeline, not an artifact of moving
$\Khat$ through the whole posterior via $\alpha$.

The third control varies the export map, one knob at a time, on the canonical
$\alpha{=}100$ exports with the fits untouched. Replacing binary opacity with a soft
count-based map ($o_k = \log(1+N_k)/\log(1+\max_j N_j)$ on the unchanged live mask,
clipped below at $10^{-3}$) gains $+0.78$ to $+4.17$\,dB (mean $+1.71$). Replacing the
scale modifier $1.41$, inherited unchanged from the VBGS render protocol, with $1.0$ gains
$+1.37$ to $+2.73$\,dB (mean $+2.09$); moving to $2.0$ instead loses PSNR on every scene
(mean $-2.18$\,dB). The two knobs were varied separately, not jointly, and VBGS was not
re-exported under alternative maps, since $1.41$ is the value its own protocol prescribes;
these deltas measure what the fixed inherited map costs the pruned DP fits, not a tuned
head-to-head. The decomposition then reads: the weight prior is exonerated, added capacity
is confirmed harmful under this pipeline, and the export map is a measured multi-dB
contributor to the deficit --- while the residual mechanism, component placement and
covariance shape under heavily pruned fits with no image-space objective anywhere in the
training loop, remains open.

\section{Limitations and discussion}
\label{sec:discussion}

\paragraph{Scope of the guarantees.} All theory concerns the mixture likelihood on colored
points. The rasterized rendering map (occlusion, alpha compositing) is applied downstream
of inference exactly as in VBGS and carries no statistical guarantees; likewise our scene
experiments use RGB-D/depth-lifted points and inherit VBGS's depth dependency. The color
model is view-independent: each component carries a single conjugate Gaussian color factor
(Eq.~\ref{eq:datamodel}), with no spherical-harmonic expansion and no dependence on viewing
direction. View-dependent appearance such as specularity is therefore outside the model
class rather than merely unrecovered by our inference, which is a real restriction relative
to standard 3DGS pipelines that encode radiance as a function of view direction. Extending
the conjugate treatment to view-dependent color, where conjugacy would have to be preserved
across a directional basis, is left to future work.

\paragraph{What the prior can and cannot do under CAVI.} Our central empirical message cuts
both ways. The DP prior gives genuine, closed-form complexity adaptation, but the realized
$\Khat$ is shaped as much by coordinate-ascent dynamics (component death without revival;
no merge moves) as by the prior, and hyperparameter learning follows the mode rather than
fixing it. We chose to implement the plain scheme and measure these effects rather than
patch them with the reassignment heuristics our approach is meant to replace; the
ELBO-guarded reordering and revival diagnostics of \S\ref{sec:khatsens} bound how much of
the under-recovery such moves recover in place, and principled revival (restricted
re-initialization between CAVI runs, merge/split proposals, or MFM-style inference,
\citealp{miller2018mixture}) is the natural next step, with our results as its
quantitative target.

\paragraph{Learning $\alpha$ is not complexity selection.} An explicit limitation:
under this mean-field family the concentration update adapts $q(\alpha)$ to the
allocation the fit is already in, so learned $\alpha$ follows the current mode rather
than steering the fit. The $\E[\alpha]$ trajectories in Figure~\ref{fig:alphatraj} show
this directly: across a two-decade spread of initializations, every run ends within one
order of magnitude of its start, $\Khat$ is set by the initialization regime rather than by
$K_{\mathrm{true}}$ (at $K_{\mathrm{true}}{=}30$: $\Khat$ $3$--$6$, $19$--$20$, $35$--$39$
from $\E[\alpha]_0 = 0.1, 1, 10$), and even where $\E[\alpha]$ drifts toward the
data-appropriate regime the allocation does not follow, having locked in earlier.
Learned $\alpha$ is within-basin adaptation, not data-driven selection of the complexity
regime; users who need a specific regime should set $\alpha$, and \S\ref{sec:khatsens}
reports what depends on that choice.

\paragraph{Rendering-grade evaluation.} We deliberately evaluated with the reference
implementation's own renderer on images and a rasterizer-free point-level metric on scenes,
so that every comparison isolates the weight prior. Full-resolution rasterized novel-view
benchmarks (PSNR/SSIM/LPIPS and the quality-versus-$\Khat$ frontier) are reported in
\S\ref{sec:raster} on all eight scenes, on evaluation views strictly disjoint from the
training frames: at the a-priori $\alpha{=}100$ setting the DP model renders
$3.40$--$9.03$\,dB below the unpruned fixed-$K$ VBGS baseline (mean $5.16$\,dB) even
though its point-space color prediction is at parity. We do not explain that deficit by the
component budget. Within each scene, over the
range of $\Khat$ the $\alpha$ sweep reaches, rendered quality is flat to declining in $\Khat$
on every scene and on all three metrics, and the best-rendering arm is never the
largest-$\alpha$ arm, so spending more components does not move the model toward VBGS: the
sweep lends no support to a component-budget account of the gap. The controlled
decomposition closing \S\ref{sec:raster} then takes the sweep's own confounds apart: a
fixed-$K$ Dirichlet control at the matched capacity renders within $+0.35$\,dB of the DP
fit while trailing VBGS by nearly the same margin, so the deficit is not the weight
prior's; a direct capacity sweep at fixed $K$ reproduces the negative quality--capacity
relation without touching $\alpha$; and varying the export map inherited from the VBGS
protocol (binary opacity to a soft count-based map, scale modifier $1.41$ to $1.0$)
recovers $1.71$ and $2.09$\,dB on average, so the fixed export map is a measured multi-dB
contributor. The residual mechanism, component placement and covariance under pruned fits
with no image-space objective, remains open, and we do not assert a substitute.

\paragraph{Uncertainty is evaluated in point space only.} One evaluation the
paper does not contain is a calibration of predicted uncertainty against \emph{rendered}
per-pixel error. Every calibration, coverage and predictive-likelihood number we report is
point-space, because $\mathrm{Var}[c\,|\,s]$ is a point-space quantity produced by the NIW
color factor, while the rasterizer is a deterministic downstream map for which no
per-pixel predictive variance is defined by the model. Producing one would require a new
construction rather than a new measurement, for instance propagating per-component color
variance through the alpha-composite weights and then testing the result against per-pixel
rendered error by sparsification or an AUSE-style curve. We have not built or tested that
construction, so no claim about rendered-error calibration appears anywhere in this paper,
and the calibration results of \S\ref{sec:uncertainty8} should not be read as evidence
about rendered images.

\paragraph{Numerical practice.} Exactness claims hold in float64. Two float64-critical
paths are documented in the repository (digamma/Beta computations; the rank-one recovery of
$\Psi$ from NIW naturals in the SVI blend), along with a scale-aware eps-level ridge used
only inside Cholesky factorizations; none affect reported results.

\paragraph{Outlook.} Because the entire pipeline stays conjugate, the door is open to
hierarchical extensions (objects as groups of components via nested constructions) and to
uncertainty-driven acquisition (next-best-view from $\mathrm{Var}[c\,|\,s]$), both of which
we leave to future work.

\subsubsection*{Broader Impact Statement}
This work is foundational statistical methodology for 3D scene representation. Calibrated
uncertainty in reconstructions can \emph{reduce} downstream harm (e.g., flagging
under-observed geometry before decisions are made on it); we are not aware of significant
risks of harm specific to this method beyond those generic to 3D reconstruction.

\subsubsection*{Reproducibility Statement}
All results are produced by config-driven scripts (one per figure) with fixed seeds; every
number in the tables is written to a JSON record by the producing script, and a repository
document maps each figure to its exact command. The verification layers are summarized in
the experiments section and Appendix~\ref{app:elbo}. The rasterized evaluation of
\S\ref{sec:raster} is the one stage that does not run on the development machine: its
fitting and export step is seeded and CPU-only, but the render-and-score step requires a
CUDA rasterizer and was executed on a rented consumer GPU whose package versions were not
pinned, so those metrics reproduce in protocol rather than bit-for-bit. Both scripts and
both commands are in the repository, and the qualification is recorded there. Code, reproduction commands, and
all experiment records are available at \url{https://github.com/archiedong/dp-splat}
under the MIT license.

\subsubsection*{Acknowledgments}
We thank the action editor and the three anonymous reviewers for a careful reading and for
requests that substantially strengthened the experimental evaluation. The Software used in
this research was created by VERSES, Inc.\ \copyright~2024 VERSES AI, Inc.

\bibliography{references}

\begin{thebibliography}{36}
\providecommand{\natexlab}[1]{#1}
\providecommand{\url}[1]{\texttt{#1}}
\expandafter\ifx\csname urlstyle\endcsname\relax
  \providecommand{\doi}[1]{doi: #1}\else
  \providecommand{\doi}{doi: \begingroup \urlstyle{rm}\Url}\fi

\bibitem[Amari(1998)]{amari1998natural}
Shun-ichi Amari.
\newblock Natural gradient works efficiently in learning.
\newblock \emph{Neural Computation}, 10\penalty0 (2):\penalty0 251--276, 1998.
\newblock \doi{10.1162/089976698300017746}.

\bibitem[Bertsekas(1999)]{bertsekas1999nonlinear}
Dimitri~P. Bertsekas.
\newblock \emph{Nonlinear Programming}.
\newblock Athena Scientific, Belmont, MA, 2 edition, 1999.

\bibitem[Bishop(2006)]{bishop2006pattern}
Christopher~M. Bishop.
\newblock \emph{Pattern Recognition and Machine Learning}.
\newblock Information Science and Statistics. Springer, New York, 2006.
\newblock ISBN 978-0-387-31073-2.

\bibitem[Blei \& Jordan(2006)Blei and Jordan]{blei2006variational}
David~M. Blei and Michael~I. Jordan.
\newblock Variational inference for {D}irichlet process mixtures.
\newblock \emph{Bayesian Analysis}, 1\penalty0 (1):\penalty0 121--143, 2006.
\newblock \doi{10.1214/06-BA104}.

\bibitem[Blei et~al.(2017)Blei, Kucukelbir, and McAuliffe]{blei2017variational}
David~M. Blei, Alp Kucukelbir, and Jon~D. McAuliffe.
\newblock Variational inference: A review for statisticians.
\newblock \emph{Journal of the American Statistical Association}, 112\penalty0
  (518):\penalty0 859--877, 2017.
\newblock \doi{10.1080/01621459.2017.1285773}.

\bibitem[Ding(2016)]{ding2016conditional}
Peng Ding.
\newblock On the conditional distribution of the multivariate $t$ distribution.
\newblock \emph{The American Statistician}, 70\penalty0 (3):\penalty0 293--295,
  2016.

\bibitem[Fan et~al.(2024)Fan, Wang, Wen, Zhu, Xu, and
  Wang]{fan2024lightgaussian}
Zhiwen Fan, Kevin Wang, Kairun Wen, Zehao Zhu, Dejia Xu, and Zhangyang Wang.
\newblock {LightGaussian}: Unbounded {3D} {G}aussian compression with 15x
  reduction and 200+ {FPS}.
\newblock In \emph{Advances in Neural Information Processing Systems},
  volume~37, pp.\  140138--140158. Curran Associates, Inc., 2024.
\newblock \doi{10.52202/079017-4447}.

\bibitem[Ferguson(1973)]{ferguson1973bayesian}
Thomas~S. Ferguson.
\newblock A {B}ayesian analysis of some nonparametric problems.
\newblock \emph{The Annals of Statistics}, 1\penalty0 (2):\penalty0 209--230,
  1973.
\newblock \doi{10.1214/aos/1176342360}.

\bibitem[Fr{\"u}hwirth-Schnatter et~al.(2021)Fr{\"u}hwirth-Schnatter,
  Malsiner-Walli, and Gr{\"u}n]{fruhwirthschnatter2021generalized}
Sylvia Fr{\"u}hwirth-Schnatter, Gertraud Malsiner-Walli, and Bettina Gr{\"u}n.
\newblock Generalized mixtures of finite mixtures and telescoping sampling.
\newblock \emph{Bayesian Analysis}, 16\penalty0 (4):\penalty0 1279--1307, 2021.
\newblock \doi{10.1214/21-BA1294}.

\bibitem[Ghahramani \& Beal(2000)Ghahramani and
  Beal]{ghahramani2000variational}
Zoubin Ghahramani and Matthew~J. Beal.
\newblock Variational inference for {B}ayesian mixtures of factor analysers.
\newblock In \emph{Advances in Neural Information Processing Systems 12}, pp.\
  449--455. MIT Press, 2000.

\bibitem[Hanson et~al.(2025)Hanson, Tu, Singla, Jayawardhana, Zwicker, and
  Goldstein]{hanson2025pup}
Alex Hanson, Allen Tu, Vasu Singla, Mayuka Jayawardhana, Matthias Zwicker, and
  Tom Goldstein.
\newblock {PUP 3D-GS}: Principled uncertainty pruning for {3D} {G}aussian
  splatting.
\newblock In \emph{Proceedings of the IEEE/CVF Conference on Computer Vision
  and Pattern Recognition (CVPR)}, pp.\  5949--5958, 2025.

\bibitem[Hoffman et~al.(2013)Hoffman, Blei, Wang, and
  Paisley]{hoffman2013stochastic}
Matthew~D. Hoffman, David~M. Blei, Chong Wang, and John Paisley.
\newblock Stochastic variational inference.
\newblock \emph{Journal of Machine Learning Research}, 14\penalty0
  (40):\penalty0 1303--1347, 2013.

\bibitem[Ishwaran \& James(2001)Ishwaran and James]{ishwaran2001gibbs}
Hemant Ishwaran and Lancelot~F. James.
\newblock Gibbs sampling methods for stick-breaking priors.
\newblock \emph{Journal of the American Statistical Association}, 96\penalty0
  (453):\penalty0 161--173, 2001.
\newblock \doi{10.1198/016214501750332758}.

\bibitem[Jiang et~al.(2024)Jiang, Lei, and Daniilidis]{jiang2024fisherrf}
Wen Jiang, Boshu Lei, and Kostas Daniilidis.
\newblock {FisherRF}: Active view selection and mapping with radiance fields
  using {F}isher information.
\newblock In \emph{Computer Vision -- ECCV 2024}, Lecture Notes in Computer
  Science, pp.\  422--440. Springer, 2024.

\bibitem[Kerbl et~al.(2023)Kerbl, Kopanas, Leimk{\"u}hler, and
  Drettakis]{kerbl2023gaussian}
Bernhard Kerbl, Georgios Kopanas, Thomas Leimk{\"u}hler, and George Drettakis.
\newblock {3D} {G}aussian splatting for real-time radiance field rendering.
\newblock \emph{ACM Transactions on Graphics}, 42\penalty0 (4):\penalty0 1--14,
  July 2023.
\newblock \doi{10.1145/3592433}.

\bibitem[Kheradmand et~al.(2024)Kheradmand, Rebain, Sharma, Sun, Tseng, Isack,
  Kar, Tagliasacchi, and Yi]{kheradmand2024mcmc}
Shakiba Kheradmand, Daniel Rebain, Gopal Sharma, Weiwei Sun, Yang-Che Tseng,
  Hossam Isack, Abhishek Kar, Andrea Tagliasacchi, and Kwang~Moo Yi.
\newblock {3D} {G}aussian splatting as {M}arkov chain {M}onte {C}arlo.
\newblock In \emph{Advances in Neural Information Processing Systems},
  volume~37, pp.\  80965--80986. Curran Associates, Inc., 2024.
\newblock \doi{10.52202/079017-2573}.

\bibitem[Kuleshov et~al.(2018)Kuleshov, Fenner, and
  Ermon]{kuleshov2018accurate}
Volodymyr Kuleshov, Nathan Fenner, and Stefano Ermon.
\newblock Accurate uncertainties for deep learning using calibrated regression.
\newblock In \emph{Proceedings of the 35th International Conference on Machine
  Learning}, volume~80 of \emph{Proceedings of Machine Learning Research}, pp.\
   2796--2804. PMLR, 2018.

\bibitem[Kurihara et~al.(2007)Kurihara, Welling, and
  Teh]{kurihara2007collapsed}
Kenichi Kurihara, Max Welling, and Yee~Whye Teh.
\newblock Collapsed variational {D}irichlet process mixture models.
\newblock In \emph{Proceedings of the Twentieth International Joint Conference
  on Artificial Intelligence ({IJCAI}-07)}, pp.\  2796--2801, 2007.

\bibitem[Le \& Yang(2015)Le and Yang]{le2015tiny}
Ya~Le and Xuan Yang.
\newblock Tiny {ImageNet} visual recognition challenge.
\newblock Technical report, Stanford University, CS231N course report, 2015.

\bibitem[Lee et~al.(2024)Lee, Rho, Sun, Ko, and Park]{lee2024compact}
Joo~Chan Lee, Daniel Rho, Xiangyu Sun, Jong~Hwan Ko, and Eunbyung Park.
\newblock Compact {3D} {G}aussian representation for radiance field.
\newblock In \emph{Proceedings of the IEEE/CVF Conference on Computer Vision
  and Pattern Recognition (CVPR)}, pp.\  21719--21728, 2024.

\bibitem[Malsiner-Walli et~al.(2016)Malsiner-Walli, Fr{\"u}hwirth-Schnatter,
  and Gr{\"u}n]{malsinerwalli2016model}
Gertraud Malsiner-Walli, Sylvia Fr{\"u}hwirth-Schnatter, and Bettina Gr{\"u}n.
\newblock Model-based clustering based on sparse finite {G}aussian mixtures.
\newblock \emph{Statistics and Computing}, 26\penalty0 (1-2):\penalty0
  303--324, 2016.
\newblock \doi{10.1007/s11222-014-9500-2}.

\bibitem[Mildenhall et~al.(2020)Mildenhall, Srinivasan, Tancik, Barron,
  Ramamoorthi, and Ng]{mildenhall2020nerf}
Ben Mildenhall, Pratul~P. Srinivasan, Matthew Tancik, Jonathan~T. Barron, Ravi
  Ramamoorthi, and Ren Ng.
\newblock {NeRF}: Representing scenes as neural radiance fields for view
  synthesis.
\newblock In Andrea Vedaldi, Horst Bischof, Thomas Brox, and Jan-Michael Frahm
  (eds.), \emph{Computer Vision -- ECCV 2020}, volume 12346 of \emph{Lecture
  Notes in Computer Science}, pp.\  405--421. Springer, 2020.
\newblock \doi{10.1007/978-3-030-58452-8_24}.

\bibitem[Miller \& Harrison(2013)Miller and Harrison]{miller2013simple}
Jeffrey~W. Miller and Matthew~T. Harrison.
\newblock A simple example of {D}irichlet process mixture inconsistency for the
  number of components.
\newblock In \emph{Advances in Neural Information Processing Systems},
  volume~26. Curran Associates, Inc., 2013.

\bibitem[Miller \& Harrison(2014)Miller and Harrison]{miller2014inconsistency}
Jeffrey~W. Miller and Matthew~T. Harrison.
\newblock Inconsistency of {P}itman--{Y}or process mixtures for the number of
  components.
\newblock \emph{Journal of Machine Learning Research}, 15:\penalty0 3333--3370,
  2014.

\bibitem[Miller \& Harrison(2018)Miller and Harrison]{miller2018mixture}
Jeffrey~W. Miller and Matthew~T. Harrison.
\newblock Mixture models with a prior on the number of components.
\newblock \emph{Journal of the American Statistical Association}, 113\penalty0
  (521):\penalty0 340--356, 2018.
\newblock \doi{10.1080/01621459.2016.1255636}.

\bibitem[Murphy(2012)]{murphy2012machine}
Kevin~P. Murphy.
\newblock \emph{Machine Learning: A Probabilistic Perspective}.
\newblock MIT Press, Cambridge, MA, 2012.
\newblock ISBN 978-0-262-01802-9.

\bibitem[Nguyen(2013)]{nguyen2013convergence}
XuanLong Nguyen.
\newblock Convergence of latent mixing measures in finite and infinite mixture
  models.
\newblock \emph{The Annals of Statistics}, 41\penalty0 (1):\penalty0 370--400,
  2013.
\newblock \doi{10.1214/12-AOS1065}.

\bibitem[Onoue et~al.(2026)Onoue, Kubo, and Ikeda]{onoue2026closedform}
Keisuke Onoue, Takatomi Kubo, and Kazushi Ikeda.
\newblock Closed-form coordinate ascent variational inference for {S}tudent-t
  process regression with {S}tudent-t likelihood.
\newblock In \emph{Proceedings of the 29th International Conference on
  Artificial Intelligence and Statistics ({AISTATS})}, 2026.
\newblock URL \url{https://openreview.net/forum?id=PN83VEfbQI}.

\bibitem[Pateux et~al.(2025)Pateux, Gendrin, Morin, Ladune, and
  Jiang]{pateux2025bogauss}
St{\'e}phane Pateux, Matthieu Gendrin, Luce Morin, Th{\'e}o Ladune, and Xiaoran
  Jiang.
\newblock {BOGausS}: Better optimized {G}aussian splatting.
\newblock \emph{arXiv preprint arXiv:2504.01844}, 2025.

\bibitem[Roth(2013)]{roth2013multivariate}
Michael Roth.
\newblock On the multivariate $t$ distribution.
\newblock Technical Report LiTH-ISY-R-3059, Link{\"o}ping University, Division
  of Automatic Control, 2013.

\bibitem[Rousseau \& Mengersen(2011)Rousseau and
  Mengersen]{rousseau2011asymptotic}
Judith Rousseau and Kerrie Mengersen.
\newblock Asymptotic behaviour of the posterior distribution in overfitted
  mixture models.
\newblock \emph{Journal of the Royal Statistical Society: Series B (Statistical
  Methodology)}, 73\penalty0 (5):\penalty0 689--710, 2011.
\newblock \doi{10.1111/j.1467-9868.2011.00781.x}.

\bibitem[Sethuraman(1994)]{sethuraman1994constructive}
Jayaram Sethuraman.
\newblock A constructive definition of {D}irichlet priors.
\newblock \emph{Statistica Sinica}, 4\penalty0 (2):\penalty0 639--650, 1994.

\bibitem[Van~de Maele et~al.(2024)Van~de Maele, {\c{C}}atal, Tschantz, Buckley,
  and Verbelen]{vandemaele2024vbgs}
Toon Van~de Maele, Ozan {\c{C}}atal, Alexander Tschantz, Christopher~L.
  Buckley, and Tim Verbelen.
\newblock Variational {B}ayes {G}aussian splatting.
\newblock \emph{arXiv preprint arXiv:2410.03592}, 2024.

\bibitem[Xu \& Zeng(2024)Xu and Zeng]{xu2024sparsevstp}
Jian Xu and Delu Zeng.
\newblock Sparse variational {S}tudent-t processes.
\newblock In \emph{Proceedings of the AAAI Conference on Artificial
  Intelligence ({AAAI})}, volume~38, pp.\  16156--16163, 2024.

\bibitem[Xu et~al.(2026)Xu, Zeng, and Paisley]{xu2026sparsevstpnn}
Jian Xu, Delu Zeng, and John Paisley.
\newblock Sparse variational {S}tudent-t processes for heavy-tailed modeling.
\newblock \emph{IEEE Transactions on Neural Networks and Learning Systems},
  2026.

\bibitem[Zhu et~al.(2026)Zhu, Zhang, Xu, and Ren]{zhu2026vbgsslam}
Yuhan Zhu, Yanyu Zhang, Jie Xu, and Wei Ren.
\newblock {VBGS-SLAM}: Variational {B}ayesian {G}aussian splatting simultaneous
  localization and mapping.
\newblock \emph{arXiv preprint arXiv:2604.02696}, 2026.

\end{thebibliography}
\bibliographystyle{tmlr}

\appendix
\section{ELBO decomposition and verification}
\label{app:elbo}

\begin{figure}[h]
\centering
\includegraphics[width=0.82\linewidth]{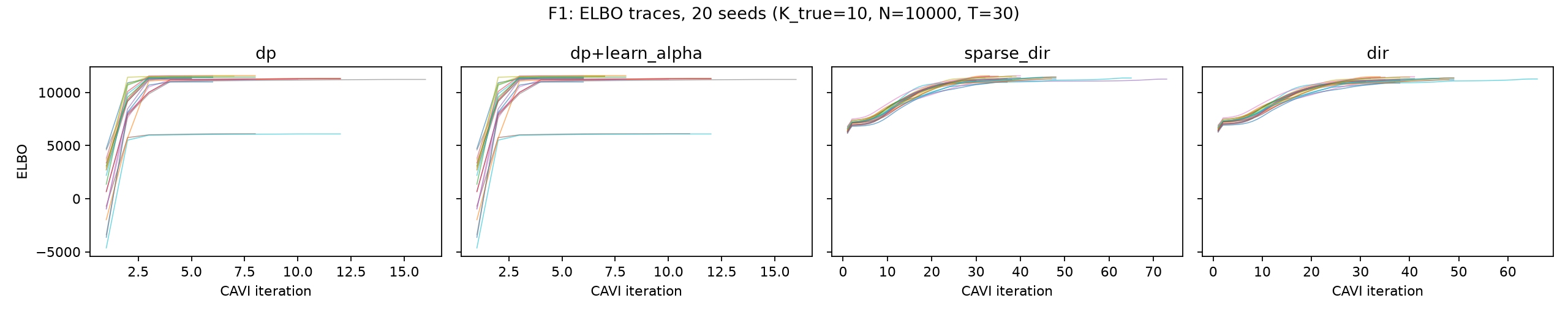}
\caption{ELBO traces, 20 seeds per panel. All traces are monotone (worst relative decrease
$-9.7\times10^{-16}$). DP variants converge $3$--$5\times$ faster in iteration count;
$\mathrm{dp(learn)}$ (labeled dp+learn\_alpha in the legend) is visually identical to
$\mathrm{dp}$. Two of the 20 DP seeds per panel ($10\%$; $95\%$ Wilson interval
$3$--$30\%$) converge to a lower local optimum.}
\label{fig:f1}
\end{figure}

The ELBO is
$\elbo = \E[\log p(X|Z,\theta)] + \E[\log p(Z|v)] + \E[\log p(v|\alpha)] +
\E[\log p(\theta)] + \E[\log p(\alpha)] - \E[\log q(Z, v, \theta, \alpha)]$,
implemented one named function per term. The NIW terms use the Wishart (precision)
parameterization $p(\Lambda) = \mathcal{W}(\Psi_0^{-1}, \nu_0)$; per-term values differ
from the Inverse-Wishart statement of \S\ref{sec:model} by the change-of-variables Jacobian
$(D{+}1)\E[\log|\Lambda_k|]$, which cancels between $\E[\log p(\theta)]$ and
$\E[\log q(\theta)]$, leaving $\elbo$ invariant (verified numerically). Every expectation
is tested against Monte Carlo draws from $q$ (Wishart/Beta/Gamma/Dirichlet sampling; $10^5$
samples for scalar terms; agreement within $5$ standard errors), and the assembled cycle is
tested for monotonicity over $200$ iterations $\times$ $20$ seeds $\times$ $4$ variants.
Figure~\ref{fig:f1} shows the corresponding ELBO traces.

\section{Summary of validation criteria}
\label{app:acceptance}

\begin{table}[h]
\centering
\caption{Validation criteria, fixed before the experiments were run, and their outcomes.}
\label{tab:acceptance}
\begin{tabular}{p{7.6cm}p{5.6cm}}
\toprule
Criterion & Outcome \\
\midrule
ELBO monotone under full-batch CAVI (200 it.\ $\times$ 20 seeds, all variants) &
Pass; worst relative decrease $-9.7\times10^{-16}$ \\
Updates match independent brute-force oracle ($N \le 200$, $T \le 10$) &
Pass at $10^{-7}$--$10^{-12}$ \\
ELBO terms match MC from $q$ & Pass within $5$ SE \\
SVI with $|B|{=}N$, $\rho_t{=}1$ reproduces CAVI & Pass at $10^{-8}$--$10^{-9}$ \\
One CAVI step reproduces reference VBGS step (baseline config) & Pass at $10^{-6}$--$10^{-8}$ \\
$\Khat$ recovers $K_{\mathrm{true}} \pm 1$ at $T = 3K_{\mathrm{true}}$ &
Pass with regime-appropriate $\alpha$ (\S\ref{sec:experiments}) \\
Image PSNR vs.\ fixed-$K$ VBGS at matched effective budget within $0.5$\,dB &
Pass, 24/24 image-$\alpha$ cells (exceeds by $+0.5$ to $+5.5$\,dB); ties converged
fixed-$K$ CAVI at the same $\Khat$ ($-0.8$ to $+0.8$\,dB) \\
\bottomrule
\end{tabular}
\end{table}

\section{Scene fitting panels}
\label{app:chair}

\begin{figure}[h]
\centering
\includegraphics[width=0.92\linewidth]{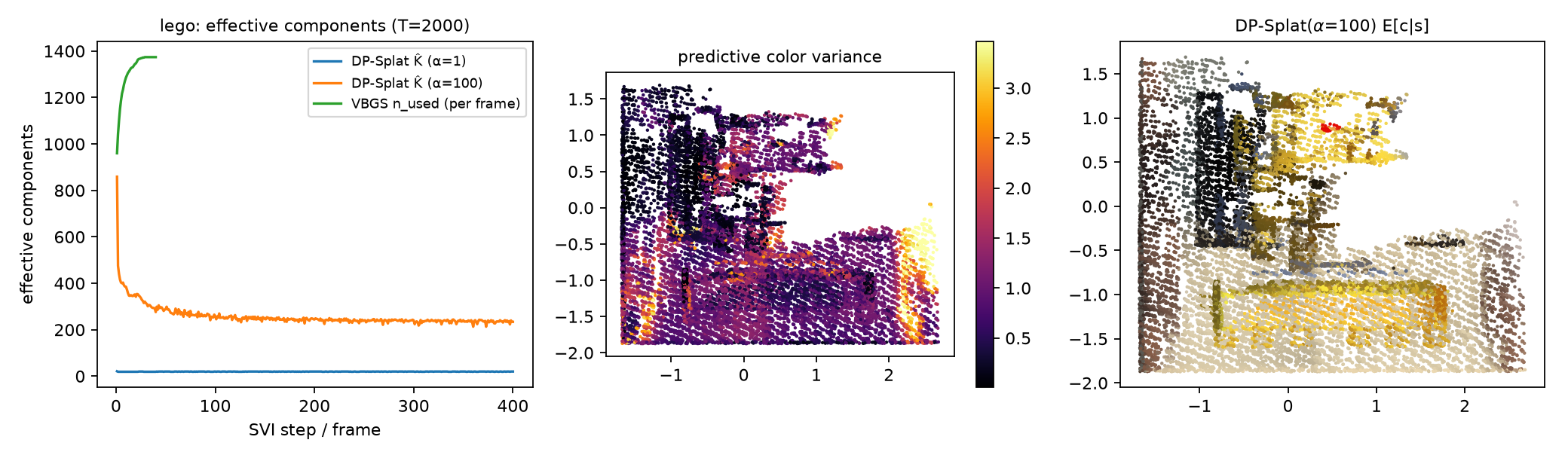}
\caption{Scene fitting (\emph{lego}). Left: effective components during continual fitting
(matched budget $T{=}2000$). The plotted DP-Splat curves are the running
single-minibatch estimate tracked during SVI; the reported $\Khat$ is recomputed from a
full-data E-step at the final posterior, and is ${\sim}260$ at $\alpha{=}100$ and
${\sim}20$ at $\alpha{=}1$ (the $\alpha{=}100$ run starts near ${\sim}850$ and sheds
redundant components). VBGS pins ${\sim}1374$ components of its budget under its reference
implementation's occupancy rule, $\Nk > 0$, or ${\sim}1292$ under the $\Nk > 1$ rule used
for $\Khat$ throughout this paper. Middle: held-out-view predictive color variance; uncertainty
concentrates on edges and occlusion boundaries. Right: held-out $\E[c|s]$ (denormalized).}
\label{fig:phase3}
\end{figure}

\begin{figure}[h]
\centering
\includegraphics[width=0.92\linewidth]{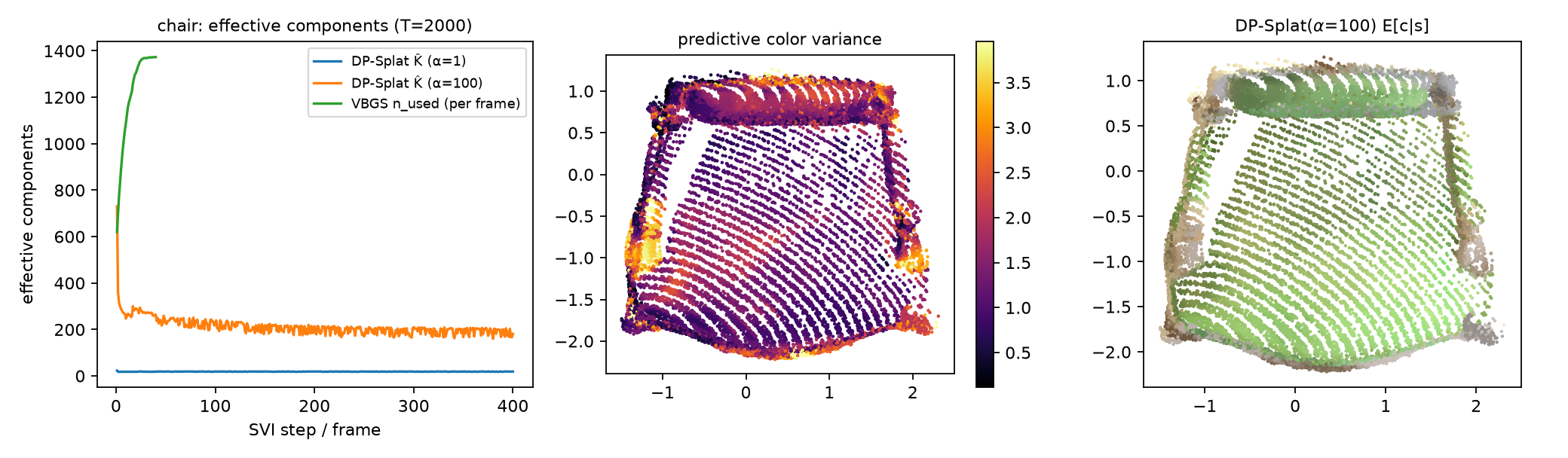}
\caption{Scene fitting (\emph{chair}), companion to Figure~\ref{fig:phase3}: VBGS pins
${\sim}1373$ components under its own $\Nk > 0$ rule (${\sim}1291$ under $\Nk > 1$);
DP-Splat self-selects ${\sim}230$ ($\alpha{=}100$) or ${\sim}19$
($\alpha{=}1$) on a full-data E-step.}
\end{figure}

\section{Learned-concentration trajectories}
\label{app:alphatraj}

\begin{figure}[h]
\centering
\includegraphics[width=\linewidth]{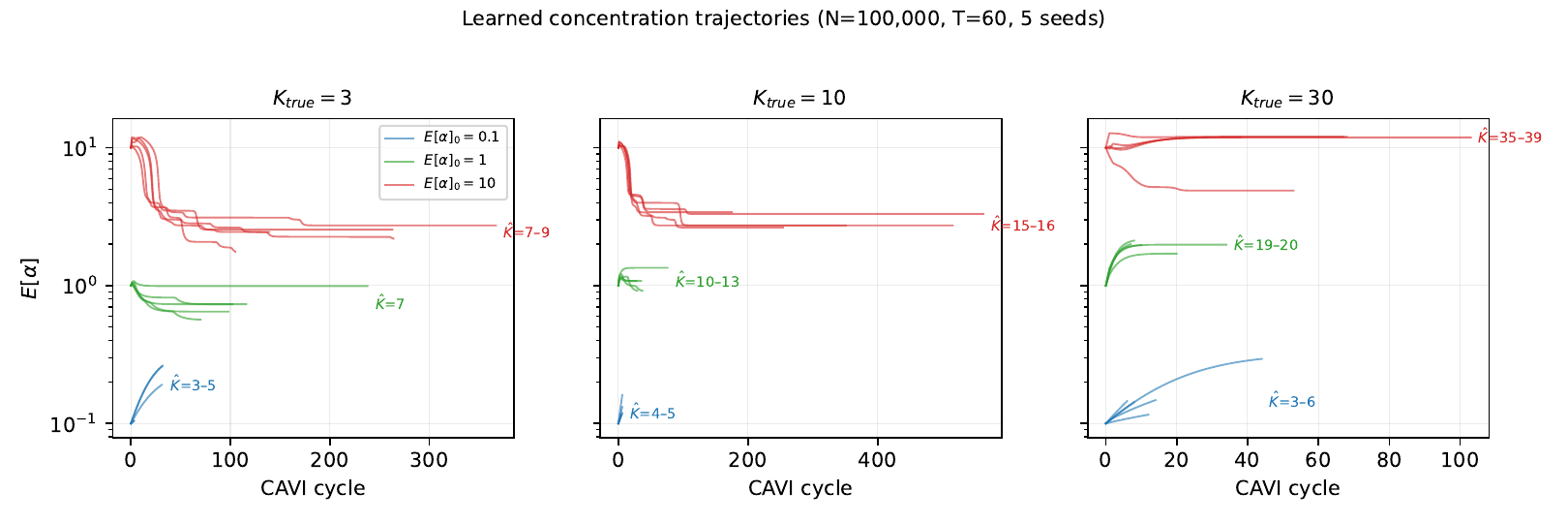}
\caption{$\E[\alpha]$ per CAVI cycle under \texttt{learn\_alpha} (log scale; one line per
seed; colors distinguish the initialization $\E[\alpha]_0$; end-of-line annotations give
the $\Khat$ range reached from that initialization). All $45$ runs converged to the
relative-ELBO tolerance under a $2{,}000$-cycle limit, the longest taking $563$ cycles.
Final $\E[\alpha]$ stays within one
order of magnitude of its initialization in every run (largest observed change a factor of
$5.7$) against the two-orders-of-magnitude spread of initializations, and $\Khat$ follows
the initialization regime rather than $K_{\mathrm{true}}$: at $K_{\mathrm{true}}{=}30$,
$\Khat$ ends at $3$--$6$, $19$--$20$, and $35$--$39$ from $\E[\alpha]_0 = 0.1, 1, 10$.
Where $\E[\alpha]$ does move toward the data-appropriate regime
($K_{\mathrm{true}}{=}3$ from $\E[\alpha]_0{=}10$ falls to $1.76$--$2.73$), $\Khat$
does not follow ($7$--$9$ vs.\ $3$): the allocation locks in early and $q(\alpha)$
then adapts to it.}
\label{fig:alphatraj}
\end{figure}

Figure~\ref{fig:alphatraj} documents the mode-following behavior discussed in
\S\ref{sec:experiments} and \S\ref{sec:discussion} at the trajectory level: the same CAVI
cycle as everywhere else, with $\E[\alpha] = w_1/w_2$ read out of $q(\alpha)$ after every
cycle. Runs start from $\E[\alpha]_0 \in \{0.1, 1, 10\}$ (by $a_0 = \E[\alpha]_0$,
$b_0 = 1$) on 3D synthetic mixtures with $K_{\mathrm{true}} \in \{3, 10, 30\}$
($N = 10^5$, $T = 60$, 5 seeds).

\section{Expanded uncertainty evaluation: full grid}
\label{app:uncertainty}

\begin{figure}[h]
\centering
\includegraphics[width=\linewidth]{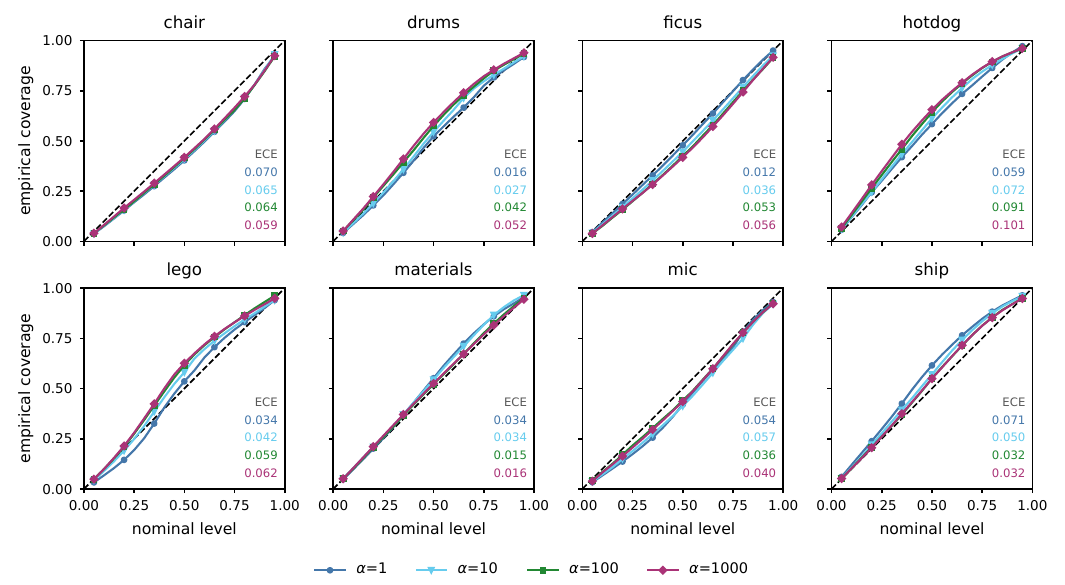}
\caption{Coverage curves (nominal vs.\ empirical central-interval coverage per color
dimension) for all eight scenes and all four $\alpha$ values, on the level grid
$0.05,0.10,\dots,0.95$; the dashed diagonal is ideal and the inset numbers are the
coverage-ECE of each curve. The upper levels are accurate everywhere: across all
$32$ scene--$\alpha$ arms the empirical coverage is within $5.3$ percentage points of
nominal at $90\%$ and within $3.6$ points at $95\%$. The visible departures sit in the
middle of the grid, and their sign is a property of the scene rather than of $\alpha$:
\emph{chair}, \emph{ficus} and \emph{mic} under-cover at every $\alpha$ (largest gap
$-10.4$ points, \emph{chair} at nominal $0.70$), while \emph{drums}, \emph{hotdog},
\emph{lego}, \emph{materials} and \emph{ship} over-cover at every $\alpha$ (largest gap
$+15.6$ points, \emph{hotdog} at nominal $0.50$). Raising $\alpha$ from $1$ to $1000$
does not move calibration in a consistent direction: it brings the $90/95\%$ levels
closer to the diagonal on \emph{drums}, \emph{hotdog}, \emph{materials} and \emph{ship}
and further from it on \emph{chair}, \emph{ficus}, \emph{lego} and \emph{mic}, and by
whole-curve coverage-ECE the split is again four scenes each way, but a different four
(improving on \emph{chair}, \emph{materials}, \emph{mic} and \emph{ship}). Calibration
here is therefore governed by the scene, not by the concentration parameter.}
\label{fig:covcurves}
\end{figure}

Table~\ref{tab:uncertaintyfull} reports the complete $\alpha$ grid behind
Table~\ref{tab:uncertainty}, and Figure~\ref{fig:covcurves} the coverage curves.

\begin{table}[h]
\centering
\small
\caption{Uncertainty evaluation on all eight NeRF-synthetic scenes, full
$\alpha \in \{1, 10, 100, 1000\}$ grid (held-out frame; point space; coverage per color
dimension; conditional held-out log-likelihood per point in normalized units).}
\label{tab:uncertaintyfull}
\begin{tabular}{llcccccc}
\toprule
Scene & $\alpha$ & $\Khat$ & PSNR & ECE & Cov.\ 68/90/95 (\%) & Cov.-ECE & Cond.\ LL/pt \\
\midrule
chair     & 1    & 19  & 14.55 & 0.0286 & 57.8/85.9/92.3 & 0.0698 & $-2.06$ \\
chair     & 10   & 81  & 15.44 & 0.0160 & 58.4/85.4/93.1 & 0.0653 & $-1.09$ \\
chair     & 100  & 230 & 16.03 & 0.0158 & 58.7/84.8/92.2 & 0.0645 & $-0.49$ \\
chair     & 1000 & 309 & 16.19 & 0.0141 & 59.4/85.1/92.5 & 0.0592 & $-0.36$ \\
\midrule
drums     & 1    & 17  & 14.36 & 0.0303 & 69.9/88.4/91.8 & 0.0165 & $+0.75$ \\
drums     & 10   & 85  & 15.41 & 0.0262 & 74.5/89.1/92.2 & 0.0268 & $+1.56$ \\
drums     & 100  & 300 & 16.29 & 0.0124 & 75.6/90.7/93.5 & 0.0420 & $+2.33$ \\
drums     & 1000 & 451 & 16.54 & 0.0105 & 76.8/91.2/93.9 & 0.0518 & $+2.43$ \\
\midrule
ficus     & 1    & 18  & 19.43 & 0.0061 & 67.2/90.5/95.0 & 0.0122 & $-1.10$ \\
ficus     & 10   & 84  & 19.52 & 0.0064 & 63.8/87.7/92.5 & 0.0364 & $-0.78$ \\
ficus     & 100  & 322 & 20.04 & 0.0075 & 61.6/85.9/91.4 & 0.0533 & $-0.28$ \\
ficus     & 1000 & 626 & 20.20 & 0.0077 & 60.8/86.1/91.6 & 0.0559 & $-0.19$ \\
\midrule
hotdog    & 1    & 18  & 19.25 & 0.0074 & 76.3/94.1/97.2 & 0.0585 & $+0.33$ \\
hotdog    & 10   & 85  & 22.49 & 0.0059 & 79.4/93.6/96.1 & 0.0723 & $+1.37$ \\
hotdog    & 100  & 261 & 24.13 & 0.0061 & 81.3/94.1/96.1 & 0.0908 & $+2.04$ \\
hotdog    & 1000 & 384 & 24.60 & 0.0059 & 81.8/94.2/96.1 & 0.1008 & $+2.32$ \\
\midrule
lego      & 1    & 20  & 14.56 & 0.0082 & 73.6/90.1/94.0 & 0.0342 & $+1.05$ \\
lego      & 10   & 85  & 16.16 & 0.0104 & 76.3/90.8/94.0 & 0.0423 & $+1.99$ \\
lego      & 100  & 260 & 17.42 & 0.0109 & 78.2/93.1/96.2 & 0.0594 & $+2.78$ \\
lego      & 1000 & 390 & 17.30 & 0.0096 & 78.4/92.0/94.9 & 0.0620 & $+2.91$ \\
\midrule
materials & 1    & 20  & 15.38 & 0.0271 & 75.7/92.4/95.5 & 0.0337 & $+0.19$ \\
materials & 10   & 94  & 15.91 & 0.0104 & 74.6/93.9/96.5 & 0.0344 & $+1.13$ \\
materials & 100  & 315 & 16.00 & 0.0048 & 70.7/91.1/95.0 & 0.0147 & $+1.57$ \\
materials & 1000 & 493 & 16.02 & 0.0042 & 70.3/90.3/94.4 & 0.0155 & $+1.70$ \\
\midrule
mic       & 1    & 19  & 16.87 & 0.0150 & 63.0/89.2/93.1 & 0.0539 & $+0.37$ \\
mic       & 10   & 69  & 17.27 & 0.0169 & 61.8/87.7/92.1 & 0.0574 & $+0.98$ \\
mic       & 100  & 164 & 17.88 & 0.0128 & 63.6/88.0/92.6 & 0.0365 & $+1.46$ \\
mic       & 1000 & 211 & 17.95 & 0.0129 & 63.7/88.1/92.2 & 0.0398 & $+1.53$ \\
\midrule
ship      & 1    & 21  & 19.15 & 0.0038 & 79.2/93.9/96.4 & 0.0711 & $-1.32$ \\
ship      & 10   & 96  & 19.85 & 0.0052 & 77.4/93.6/95.9 & 0.0497 & $-0.43$ \\
ship      & 100  & 330 & 20.04 & 0.0046 & 74.7/92.1/95.0 & 0.0320 & $-0.15$ \\
ship      & 1000 & 634 & 20.12 & 0.0033 & 74.9/92.0/94.8 & 0.0321 & $-0.04$ \\
\bottomrule
\end{tabular}
\end{table}

\section{Full $\Khat$ tables}
\label{app:tables}

Complete per-cell records (including $n_{\min}$ sensitivity and entropy-based effective
$K$) for Figures~\ref{fig:f2}, \ref{fig:khatsens}, and~\ref{fig:f4} ship as
machine-readable JSON with the code;
the included repository document \texttt{RESULTS.md} renders them as tables.

\clearpage
\section{Matched-capacity fixed-$K$ Dirichlet ablation, per scene}
\label{app:dirablation}

Table~\ref{tab:dirablation} gives the per-scene detail behind the ablation summarized
in \S\ref{sec:uncertainty8}. Each scene contributes two arms fit under an identical SVI
protocol and scored on the same held-out frame: the DP-Splat fit at the a-priori
$\alpha{=}100$, and a fixed-$K$ symmetric Dirichlet mixture with $e_0 = 1/K$ at
$K = \Khat$, the capacity the DP arm selects at that $\alpha$. Both arms are run over the
same three seeds, with every random element reseeded, including the $k$-means
initialization. Entries are seed means $\pm$ standard deviations over the three seeds. The
$K$ column is the capacity handed to the fixed-$K$ arm, read off the $\alpha{=}100$ row of
Table~\ref{tab:uncertainty}; the DP arm keeps the truncation $T = 2000$ used everywhere
else and selects its own capacity, which is why its occupancy varies from seed to seed. The
column ``occ.'' is the number of components each arm actually occupies at the final
posterior under the $\Nk > 1$ rule used for $\Khat$ throughout the paper, evaluated on a
full-data E-step; it shows that the fixed-$K$ arm uses the capacity it was given rather
than emptying it. Lower is better for both error columns, higher for the last two.

\begin{table}[h]
\centering
\small
\setlength{\tabcolsep}{4.5pt}
\caption{Matched-capacity ablation on all eight NeRF-synthetic scenes: DP-Splat at
$\alpha{=}100$ against a fixed-$K$ symmetric Dirichlet mixture at $K = \Khat$, three seeds
per arm (mean $\pm$ sd). Regression-ECE and coverage-ECE are in the units of
Table~\ref{tab:uncertaintyfull}; conditional held-out log-likelihood is per point in
normalized units; point-PSNR is in original $[0,1]$ color units. Paired over the eight
scenes, the fixed-$K$ arm differs from the DP arm by $-5\times10^{-5}$ in regression-ECE
($p = 0.86$), $-9\times10^{-5}$ in coverage-ECE ($p = 0.89$), $-0.010$ nats in conditional
log-likelihood ($p = 0.44$) and $-0.077$\,dB in point-PSNR ($p = 0.022$, which does not
clear a four-test Bonferroni bar of $0.0125$ and is smaller than the mean per-scene seed
spread of $0.121$\,dB).}
\label{tab:dirablation}
\begin{tabular}{llrrcccc}
\toprule
Scene & Arm & $K$ & occ. & ECE & Cov.-ECE & Cond.\ LL/pt & PSNR (dB) \\
\midrule
\emph{chair} & DP-Splat $\alpha{=}100$ &     & 240 & $0.0158 \pm 0.0009$ & $0.0648 \pm 0.0004$ & $-0.492 \pm 0.013$ & $16.030 \pm 0.027$ \\
             & fixed-$K$ Dirichlet     & 230 & 199 & $0.0156 \pm 0.0006$ & $0.0632 \pm 0.0012$ & $-0.530 \pm 0.015$ & $15.968 \pm 0.029$ \\
\addlinespace
\emph{drums} & DP-Splat $\alpha{=}100$ &     & 297 & $0.0125 \pm 0.0002$ & $0.0468 \pm 0.0042$ & $+2.283 \pm 0.049$ & $16.277 \pm 0.014$ \\
             & fixed-$K$ Dirichlet     & 300 & 288 & $0.0115 \pm 0.0015$ & $0.0465 \pm 0.0045$ & $+2.289 \pm 0.067$ & $16.243 \pm 0.008$ \\
\addlinespace
\emph{ficus} & DP-Splat $\alpha{=}100$ &     & 323 & $0.0075 \pm 0.0002$ & $0.0515 \pm 0.0039$ & $-0.296 \pm 0.016$ & $20.121 \pm 0.072$ \\
             & fixed-$K$ Dirichlet     & 322 & 307 & $0.0088 \pm 0.0004$ & $0.0548 \pm 0.0025$ & $-0.318 \pm 0.018$ & $20.137 \pm 0.029$ \\
\addlinespace
\emph{hotdog} & DP-Splat $\alpha{=}100$ &     & 258 & $0.0059 \pm 0.0002$ & $0.0912 \pm 0.0039$ & $+2.080 \pm 0.043$ & $24.178 \pm 0.137$ \\
             & fixed-$K$ Dirichlet     & 261 & 251 & $0.0048 \pm 0.0005$ & $0.0900 \pm 0.0057$ & $+2.139 \pm 0.028$ & $23.947 \pm 0.293$ \\
\addlinespace
\emph{lego} & DP-Splat $\alpha{=}100$ &     & 261 & $0.0108 \pm 0.0003$ & $0.0583 \pm 0.0017$ & $+2.794 \pm 0.023$ & $17.286 \pm 0.119$ \\
            & fixed-$K$ Dirichlet     & 260 & 241 & $0.0111 \pm 0.0022$ & $0.0590 \pm 0.0059$ & $+2.791 \pm 0.023$ & $17.163 \pm 0.273$ \\
\addlinespace
\emph{materials} & DP-Splat $\alpha{=}100$ &     & 309 & $0.0051 \pm 0.0006$ & $0.0108 \pm 0.0036$ & $+1.568 \pm 0.021$ & $15.917 \pm 0.097$ \\
                 & fixed-$K$ Dirichlet     & 315 & 310 & $0.0046 \pm 0.0011$ & $0.0084 \pm 0.0027$ & $+1.543 \pm 0.006$ & $15.834 \pm 0.016$ \\
\addlinespace
\emph{mic} & DP-Splat $\alpha{=}100$ &     & 168 & $0.0129 \pm 0.0005$ & $0.0380 \pm 0.0025$ & $+1.439 \pm 0.024$ & $17.895 \pm 0.024$ \\
           & fixed-$K$ Dirichlet     & 164 & 145 & $0.0131 \pm 0.0007$ & $0.0392 \pm 0.0025$ & $+1.386 \pm 0.034$ & $17.857 \pm 0.011$ \\
\addlinespace
\emph{ship} & DP-Splat $\alpha{=}100$ &     & 335 & $0.0044 \pm 0.0002$ & $0.0321 \pm 0.0018$ & $-0.156 \pm 0.030$ & $19.993 \pm 0.041$ \\
            & fixed-$K$ Dirichlet     & 330 & 316 & $0.0050 \pm 0.0007$ & $0.0316 \pm 0.0012$ & $-0.161 \pm 0.033$ & $19.937 \pm 0.077$ \\
\bottomrule
\end{tabular}
\end{table}

\clearpage
\section{Binned variance-calibration curves, all eight scenes}
\label{app:calcurves}

\begin{figure}[h]
\centering
\includegraphics[width=\linewidth]{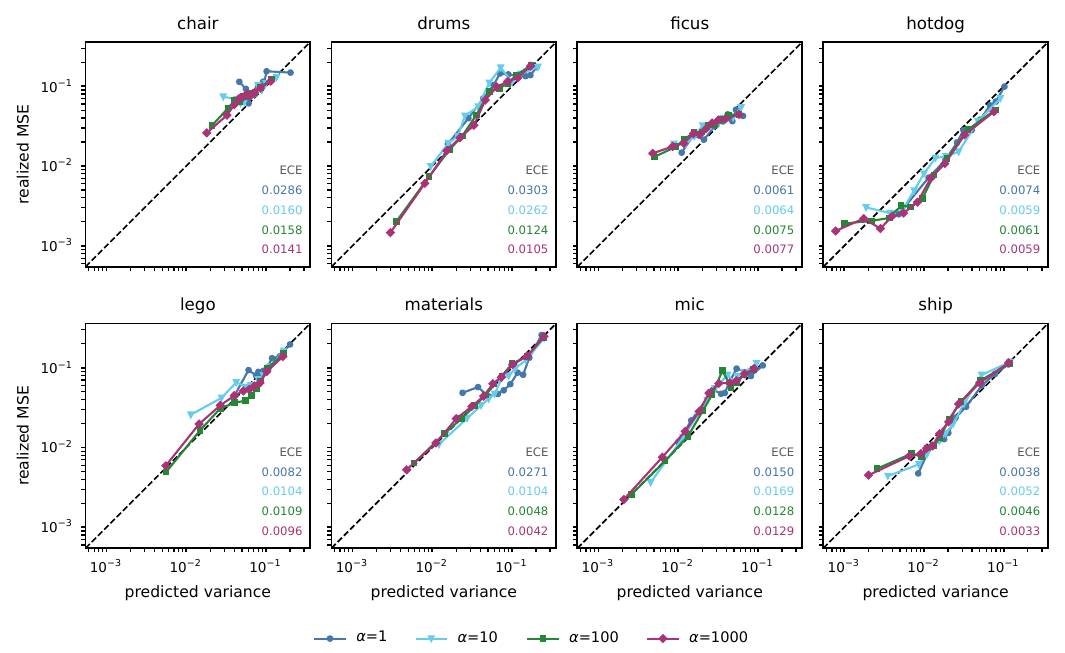}
\caption{Binned variance calibration for all eight scenes and all four $\alpha$
values: mean predicted color variance per decile bin against the realized mean squared
error of the held-out colors in that bin (log--log; the dashed diagonal is ideal; inset
numbers are the regression-ECE of each curve, in $[0,1]$ color units). The curves track the
diagonal over roughly two decades of predicted variance on every scene, and the ordering
within a scene is preserved: predicted variance and realized error are positively
correlated at the point level on all $32$ scene--$\alpha$ arms (Pearson $r$ from $0.06$ on
\emph{chair} at $\alpha{=}1$ to $0.51$ on \emph{hotdog} at $\alpha{=}10$). What
misspecification costs is the global scale rather than the shape, and its sign is a
property of the scene: at $\alpha{=}100$ the mean predicted variance over-estimates the
realized MSE on \emph{hotdog}, \emph{lego} and \emph{materials} and under-estimates it on
the other five, the largest mismatch anywhere in the grid being \emph{hotdog} at
$\alpha{=}1000$, which over-estimates by a factor of $1.55$. Regression-ECE stays between
$0.0033$ and $0.0303$ across all $32$ arms.}
\label{fig:calcurves}
\end{figure}

Figure~\ref{fig:calcurves} is the misspecified-regime counterpart of the model-matched
synthetic calibration curve of Figure~\ref{fig:f5} (left), drawn from the same decile
binning the regression-ECE of Tables~\ref{tab:uncertainty} and \ref{tab:uncertaintyfull} is
computed from, so the per-panel annotations are the numbers those tables quote. It is
placed here rather than in the main text because the scalar ECE already carries the
comparison; the curves show where in the variance range the residual miscalibration sits.

\section{Derivation of the predictive conditional moments (Eq.~\ref{eq:condmoments})}
\label{app:eq7}

\subsection*{Setup and notation}

Data points are $x = (s, c) \in \mathbb{R}^{D_s} \times \mathbb{R}^{D_c}$. The per-component likelihood factorizes across modalities,
\[
p(x \mid \theta_k) \;=\; \mathcal{N}\!\bigl(s \mid \mu_{k,s}, \Sigma_{k,s}\bigr)\,
\mathcal{N}\!\bigl(c \mid \mu_{k,c}, \Sigma_{k,c}\bigr),
\qquad \theta_k = (\theta_{k,s}, \theta_{k,c}), \quad \theta_{k,m} = (\mu_{k,m}, \Sigma_{k,m}),
\]
and the mixture likelihood is $p(x \mid \pi, \theta) = \sum_k \pi_k\, p(x \mid \theta_k)$. The variational posterior is mean-field:
\[
q(\pi, \theta) \;=\; q(\pi)\; \prod_k q(\theta_{k,s})\, q(\theta_{k,c}),
\qquad
q(\theta_{k,m}) = \mathrm{NIW}\bigl(m_{k,m}, \kappa_{k,m}, \Psi_{k,m}, \nu_{k,m}\bigr).
\]
\textbf{NIW convention.} $\mathrm{NIW}(m,\kappa,\Psi,\nu)$ means
$\Sigma \sim \mathrm{IW}(\Psi, \nu)$ with density
\[
\mathrm{IW}(\Sigma \mid \Psi, \nu)
= \frac{|\Psi|^{\nu/2}}{2^{\nu D/2}\,\Gamma_D(\nu/2)}\,
|\Sigma|^{-(\nu+D+1)/2}\, e^{-\frac{1}{2}\operatorname{tr}(\Psi \Sigma^{-1})},
\qquad \nu > D-1,
\]
and $\mu \mid \Sigma \sim \mathcal{N}(m, \Sigma/\kappa)$. Here $\Gamma_D(a) = \pi^{D(D-1)/4}\prod_{i=1}^D \Gamma\!\bigl(a + \tfrac{1-i}{2}\bigr)$. This is the convention of \citet[\S4.6.3]{murphy2012machine}; equivalently the precision satisfies $\Sigma^{-1} \sim \mathrm{Wishart}(\Psi^{-1}, \nu)$. The conjugate update in \texttt{niw.py} ($\Psi_n = \Psi_0 + S + \tfrac{\kappa_0 N}{\kappa_0+N}(\bar{x}-m_0)(\bar{x}-m_0)^{\!\top}$, $\nu_n = \nu_0 + N$) is stated in exactly this convention, so the predictive formula below applies.

The multivariate Student-$t$ density with location $\mu$, scale matrix $S \succ 0$, and degrees of freedom $\eta > 0$ is
\[
\mathrm{St}(x \mid \mu, S, \eta)
= \frac{\Gamma\!\bigl(\tfrac{\eta+D}{2}\bigr)}{\Gamma\!\bigl(\tfrac{\eta}{2}\bigr)\,(\eta\pi)^{D/2}\,|S|^{1/2}}
\Bigl[1 + \tfrac{1}{\eta}(x-\mu)^{\!\top} S^{-1} (x-\mu)\Bigr]^{-(\eta+D)/2}.
\]

\subsection*{Step 1: $\mathbb{E}_q$ of the mixture is the mixture of $\mathbb{E}_q$'s}

The posterior predictive under $q$ is $\hat{p}(x^*) := \mathbb{E}_q\bigl[p(x^* \mid \pi, \theta)\bigr]$. For fixed $x^*$,
\begin{align*}
\hat{p}(x^*)
&= \mathbb{E}_q\Bigl[\sum_k \pi_k\, \mathcal{N}(s^* \mid \theta_{k,s})\, \mathcal{N}(c^* \mid \theta_{k,c})\Bigr]
\overset{(i)}{=} \sum_k \mathbb{E}_q\bigl[\pi_k\, \mathcal{N}(s^* \mid \theta_{k,s})\, \mathcal{N}(c^* \mid \theta_{k,c})\bigr] \\
&\overset{(ii)}{=} \sum_k \mathbb{E}[\pi_k]\; \mathbb{E}\bigl[\mathcal{N}(s^* \mid \theta_{k,s})\bigr]\; \mathbb{E}\bigl[\mathcal{N}(c^* \mid \theta_{k,c})\bigr].
\end{align*}
$(i)$ is linearity of expectation; for a finite (truncated) mixture it is trivial, and for the infinite DP sum it is justified by Tonelli's theorem since every term is nonnegative. $(ii)$ holds because under the mean-field $q$, the random variables $\pi_k$, $\mathcal{N}(s^* \mid \theta_{k,s})$, and $\mathcal{N}(c^* \mid \theta_{k,c})$ are functions of the mutually independent blocks $\pi$, $\theta_{k,s}$, $\theta_{k,c}$, and the expectation of a product of independent, nonnegative, integrable random variables factorizes (Fubini). Note the $k$-th term only needs $\pi \perp \theta_{k,s} \perp \theta_{k,c}$; independence across components $k$ is not even required for this identity, though mean-field provides it. Each factor $\mathbb{E}[\mathcal{N}(\cdot \mid \theta_{k,m})]$ is evaluated in Step 2, giving \emph{exactly}
\[
\boxed{\;\hat{p}(s^*, c^*) = \sum_k \mathbb{E}[\pi_k]\; \mathrm{St}_s(s^* \mid k)\; \mathrm{St}_c(c^* \mid k).\;}
\]
This identity is pointwise-exact in $x^*$ under $q$; there is no approximation beyond the mean-field posterior itself. (Under a truncated stick-breaking $q(\pi) = \prod_k \mathrm{Beta}(v_k \mid \gamma_{k1}, \gamma_{k2})$, $\mathbb{E}[\pi_k] = \mathbb{E}[v_k]\prod_{j<k}\mathbb{E}[1 - v_j]$ by the same independence argument.)

\subsection*{Step 2: NIW $\Rightarrow$ Student-$t$ predictive}

Fix a modality and drop indices: $q(\mu, \Sigma) = \mathrm{NIW}(m, \kappa, \Psi, \nu)$, dimension $D$. We compute
$p(x) = \iint \mathcal{N}(x \mid \mu, \Sigma)\, \mathcal{N}(\mu \mid m, \Sigma/\kappa)\, \mathrm{IW}(\Sigma \mid \Psi, \nu)\, d\mu\, d\Sigma$.

\paragraph{2a: Marginalize $\mu$.} Given $\Sigma$, write $x = \mu + \varepsilon$ with $\varepsilon \sim \mathcal{N}(0, \Sigma)$ independent of $\mu \sim \mathcal{N}(m, \Sigma/\kappa)$. Sums of independent Gaussians give
\[
\int \mathcal{N}(x \mid \mu, \Sigma)\, \mathcal{N}(\mu \mid m, \Sigma/\kappa)\, d\mu
= \mathcal{N}\!\Bigl(x \,\Big|\, m,\; \tfrac{\kappa+1}{\kappa}\,\Sigma\Bigr).
\]

\paragraph{2b: Marginalize $\Sigma$.} First a lemma: if $\Sigma \sim \mathrm{IW}(\Psi, \nu)$ and $b > 0$, then $b\Sigma \sim \mathrm{IW}(b\Psi, \nu)$ (substitute $T = b\Sigma$ in the density; the Jacobian $b^{-D(D+1)/2}$ is absorbed into the normalizer and $\operatorname{tr}(\Psi(T/b)^{-1}) = \operatorname{tr}((b\Psi)T^{-1})$). Hence it suffices to evaluate, with $d := x - m$,
\[
p(x) = \int_{\Sigma \succ 0} \mathcal{N}(x \mid m, \Sigma)\, \mathrm{IW}(\Sigma \mid \tilde{\Psi}, \nu)\, d\Sigma,
\qquad \tilde{\Psi} := \tfrac{\kappa+1}{\kappa}\Psi .
\]
Using $d^{\!\top}\Sigma^{-1}d = \operatorname{tr}(dd^{\!\top}\Sigma^{-1})$, the integrand is
\[
\frac{|\tilde\Psi|^{\nu/2}}{2^{\nu D/2}\Gamma_D(\nu/2)\,(2\pi)^{D/2}}\;
|\Sigma|^{-\frac{(\nu+1)+D+1}{2}}\,
\exp\!\Bigl\{-\tfrac{1}{2}\operatorname{tr}\bigl[(\tilde\Psi + dd^{\!\top})\Sigma^{-1}\bigr]\Bigr\},
\]
i.e.\ an unnormalized $\mathrm{IW}(\tilde\Psi + dd^{\!\top},\, \nu+1)$ kernel. The IW normalization
$\int |\Sigma|^{-(\nu'+D+1)/2} e^{-\frac{1}{2}\operatorname{tr}(A\Sigma^{-1})}\,d\Sigma = 2^{\nu' D/2}\,\Gamma_D(\nu'/2)\,|A|^{-\nu'/2}$
with $\nu' = \nu+1$, $A = \tilde\Psi + dd^{\!\top}$ yields
\[
p(x) = \pi^{-D/2}\;
\frac{\Gamma_D\!\bigl(\tfrac{\nu+1}{2}\bigr)}{\Gamma_D\!\bigl(\tfrac{\nu}{2}\bigr)}\;
\frac{|\tilde\Psi|^{\nu/2}}{|\tilde\Psi + dd^{\!\top}|^{(\nu+1)/2}}
\qquad\text{(the powers of $2$ cancel: } -\tfrac{\nu D}{2} - \tfrac{D}{2} + \tfrac{(\nu+1)D}{2} = 0\text{)}.
\]
The multivariate-gamma ratio telescopes:
\[
\frac{\Gamma_D\!\bigl(\tfrac{\nu+1}{2}\bigr)}{\Gamma_D\!\bigl(\tfrac{\nu}{2}\bigr)}
= \prod_{i=1}^{D} \frac{\Gamma\!\bigl(\tfrac{\nu+2-i}{2}\bigr)}{\Gamma\!\bigl(\tfrac{\nu+1-i}{2}\bigr)}
= \frac{\Gamma\!\bigl(\tfrac{\nu+1}{2}\bigr)}{\Gamma\!\bigl(\tfrac{\nu+1-D}{2}\bigr)},
\]
and the matrix determinant lemma gives $|\tilde\Psi + dd^{\!\top}| = |\tilde\Psi|\,(1 + d^{\!\top}\tilde\Psi^{-1}d)$. Therefore
\[
p(x) = \frac{\Gamma\!\bigl(\tfrac{\nu+1}{2}\bigr)}{\Gamma\!\bigl(\tfrac{\nu+1-D}{2}\bigr)\,\pi^{D/2}\,|\tilde\Psi|^{1/2}}
\bigl(1 + d^{\!\top}\tilde\Psi^{-1}d\bigr)^{-(\nu+1)/2}.
\]

\paragraph{2c: Identification as Student-$t$.} Match against $\mathrm{St}(x \mid m, S, \eta)$: equate the tail exponents, $\eta + D = \nu + 1 \Rightarrow \eta = \nu - D + 1$, and the quadratic forms, $S^{-1}/\eta = \tilde\Psi^{-1} \Rightarrow S = \tilde\Psi/\eta$. The normalizers then agree identically: $\Gamma(\tfrac{\eta+D}{2}) = \Gamma(\tfrac{\nu+1}{2})$, $\Gamma(\tfrac{\eta}{2}) = \Gamma(\tfrac{\nu+1-D}{2})$, and $(\eta\pi)^{D/2}|S|^{1/2} = \eta^{D/2}\pi^{D/2}\,\eta^{-D/2}|\tilde\Psi|^{1/2} = \pi^{D/2}|\tilde\Psi|^{1/2}$. Substituting $\tilde\Psi = \tfrac{\kappa+1}{\kappa}\Psi$:
\[
\boxed{\;p(x) = \mathrm{St}\!\Bigl(x \,\Big|\, m,\;
\Psi\,\frac{\kappa+1}{\kappa\,(\nu - D + 1)},\;
\nu - D + 1\Bigr).\;}
\]
This is exactly \citet[eq.~(4.222)]{murphy2012machine}: $p(x \mid \mathcal{D}) = t_{\nu_N - D + 1}\bigl(m_N,\, \tfrac{\kappa_N+1}{\kappa_N(\nu_N - D + 1)} S_N\bigr)$ with $S_N \equiv \Psi_N$ the IW scale. It agrees with \citet[eq.~(10.81)]{bishop2006pattern} after translating to the Wishart-on-precision parameterization: $L_k = \tfrac{(\nu_k+1-D)\beta_k}{1+\beta_k}W_k \Leftrightarrow L_k^{-1} = \tfrac{\beta_k+1}{\beta_k(\nu_k+1-D)}\Psi_k$ with $W_k = \Psi_k^{-1}$, $\beta_k = \kappa_k$. Note $\eta = \nu - D + 1 > 0$ is automatic from IW properness ($\nu > D - 1$), so the predictive is always a proper density.

\subsection*{Step 3: Conditioning the mixture on the sub-vector $s = s^*$}

Write $a_k := \mathbb{E}[\pi_k]$, $f_k(s) := \mathrm{St}_s(s \mid k)$, $g_k(c) := \mathrm{St}_c(c \mid k)$, so $\hat p(s,c) = \sum_k a_k f_k(s) g_k(c)$ with $\sum_k a_k = 1$. Since each $g_k$ is a normalized density, the spatial marginal is $\hat p(s) = \int \hat p(s,c)\,dc = \sum_j a_j f_j(s)$, and $\hat p(s^*) > 0$ for every $s^*$ because Student-$t$ densities are strictly positive. Hence
\[
\hat p(c \mid s^*) = \frac{\hat p(s^*, c)}{\hat p(s^*)}
= \sum_k w_k(s^*)\, g_k(c),
\qquad
\boxed{\;w_k(s^*) = \frac{a_k\, f_k(s^*)}{\sum_j a_j\, f_j(s^*)} = \frac{\mathbb{E}[\pi_k]\,\mathrm{St}_s(s^* \mid k)}{\sum_j \mathbb{E}[\pi_j]\,\mathrm{St}_s(s^* \mid j)}.\;}
\]
\textbf{The critical subtlety.} The general rule for conditioning any mixture is $p(c \mid s^*) = \sum_k w_k(s^*)\, p_k(c \mid s^*)$ with the \emph{within-component} conditional $p_k(c \mid s^*) = p_k(s^*, c)/p_k(s^*)$. Here each component's joint predictive is a genuine product $p_k(s,c) = f_k(s)\,g_k(c)$: spatial and color are \emph{independent Student-$t$'s within a component}, because $q$ carries \emph{separate} NIW factors $q(\theta_{k,s})\,q(\theta_{k,c})$ (equivalently: two independent Gamma mixing variables, one per modality). Therefore $p_k(c \mid s^*) = g_k(c)$ with \emph{no} conditioning correction: no location shift, no scale inflation, no dof change.

\emph{This would NOT hold for a joint Student-$t$.} If instead $p_k(s,c) = \mathrm{St}\bigl((s,c) \mid (m_s, m_c), S, \eta\bigr)$ with blocks $S_{ss}, S_{sc}, S_{cc}$ (as would arise from a single NIW over the full $(D_s{+}D_c)$-dimensional covariance), the conditional is \citep{roth2013multivariate,ding2016conditional}
\[
c \mid s^* \;\sim\; \mathrm{St}\!\Bigl(
m_c + S_{cs}S_{ss}^{-1}(s^* - m_s),\;
\frac{\eta + \delta_s}{\eta + D_s}\bigl(S_{cc} - S_{cs}S_{ss}^{-1}S_{sc}\bigr),\;
\eta + D_s\Bigr),
\quad
\delta_s = (s^* - m_s)^{\!\top} S_{ss}^{-1} (s^* - m_s).
\]
Even with block-diagonal scale ($S_{cs} = 0$) the conditional scale still depends on $s^*$ through $\tfrac{\eta+\delta_s}{\eta+D_s}$ and the dof rises to $\eta + D_s$: for a joint $t$, uncorrelated $\neq$ independent, because both blocks share one latent Gamma mixing variable. The no-correction step is thus valid \emph{only} in the factorized (product-of-independent-$t$'s) case, which is exactly the case here.

\subsection*{Step 4: Conditional moments via the mixture law of total expectation/variance}

Introduce the latent label $z$ with $\Pr(z = k \mid s^*) = w_k(s^*)$ and $c \mid z{=}k \sim g_k = \mathrm{St}(m_{k,c}, S_{k,c}, \eta_{k,c})$. By the law of total expectation (each component mean exists iff $\eta_{k,c} > 1$, and $\mathbb{E}_{g_k}[c] = m_{k,c}$ by symmetry):
\[
\boxed{\;\mathbb{E}[c \mid s^*] = \sum_k w_k(s^*)\, m_{k,c} =: \bar{m}(s^*).\;}
\]
Second moment (each $\mathbb{E}_{g_k}[cc^{\!\top}] = C_k + m_{k,c}m_{k,c}^{\!\top}$ with $C_k := \mathrm{Cov}_{g_k}[c]$, finite iff $\eta_{k,c} > 2$; see Step 5):
\[
\mathbb{E}[cc^{\!\top} \mid s^*] = \sum_k w_k \bigl(C_k + m_{k,c}m_{k,c}^{\!\top}\bigr),
\qquad
\boxed{\;\mathrm{Cov}[c \mid s^*] = \sum_k w_k \bigl(C_k + m_{k,c}m_{k,c}^{\!\top}\bigr) - \bar m\,\bar m^{\!\top}.\;}
\]
Equivalently, in the (algebraically identical, numerically safer) law-of-total-variance form:
\[
\mathrm{Cov}[c \mid s^*] = \underbrace{\sum_k w_k\, C_k}_{\text{within-component}} \;+\; \underbrace{\sum_k w_k\, (m_{k,c} - \bar m)(m_{k,c} - \bar m)^{\!\top}}_{\text{between-component}} \;\succeq\; 0 .
\]

\subsection*{Step 5: Student-$t$ covariance factor and its domain}

Gaussian scale-mixture representation: if $u \sim \mathrm{Gamma}(\tfrac{\eta}{2}, \tfrac{\eta}{2})$ (shape, rate) and $c \mid u \sim \mathcal{N}(m, S/u)$, then marginally $c \sim \mathrm{St}(m, S, \eta)$. Verification: with $\delta = d^{\!\top}S^{-1}d$,
\[
\int_0^\infty \mathcal{N}(c \mid m, S/u)\,\mathrm{Ga}(u \mid \tfrac{\eta}{2}, \tfrac{\eta}{2})\,du
\propto \int_0^\infty u^{\frac{\eta+D}{2}-1} e^{-\frac{u}{2}(\eta + \delta)}\,du
= \Gamma\!\bigl(\tfrac{\eta+D}{2}\bigr)\Bigl(\tfrac{\eta+\delta}{2}\Bigr)^{-\frac{\eta+D}{2}}
\propto \bigl(1 + \tfrac{\delta}{\eta}\bigr)^{-\frac{\eta+D}{2}},
\]
and tracking constants reproduces the $\mathrm{St}$ normalizer exactly. Then by the law of total variance,
\[
\mathrm{Cov}[c] = \mathbb{E}\bigl[\mathrm{Cov}(c \mid u)\bigr] + \mathrm{Cov}\bigl(\mathbb{E}[c \mid u]\bigr)
= S\,\mathbb{E}[u^{-1}] + 0 .
\]
For $u \sim \mathrm{Gamma}(a, b)$, $\mathbb{E}[u^{-1}] = \int u^{-1}\tfrac{b^a}{\Gamma(a)}u^{a-1}e^{-bu}du = \tfrac{b^a\,\Gamma(a-1)}{\Gamma(a)\,b^{a-1}} = \tfrac{b}{a-1}$, finite iff $a > 1$. With $a = b = \eta/2$:
\[
\boxed{\;\mathrm{Cov}[c] = \frac{\eta}{\eta - 2}\, S, \qquad \text{finite iff } \eta > 2.\;}
\]
Domain summary: $\eta \le 1$: mean undefined; $1 < \eta \le 2$: mean $= m$ but covariance infinite; $\eta > 2$: both exist. In DP-Splat, $\eta_{k,c} = \nu_{k,c} - D_c + 1 = \nu_0 + N_k - D_c + 1$; under the default prior $\nu_0 = D_c + 2$ this is $N_k + 3 > 2$ for all $N_k \ge 0$, and \texttt{predictive.py} guards non-default priors at runtime.

\subsection*{Correspondence with Eq.~\eqref{eq:condmoments} and the implementation}

Every element of Eq.~\eqref{eq:condmoments} follows from the derivation: (a) the exact mixture-of-products predictive $\hat p = \sum_k \mathbb{E}[\pi_k]\,\mathrm{St}_s\,\mathrm{St}_c$ (Step 1); (b) $S_k = \Psi_k \tfrac{\kappa_k+1}{\kappa_k \eta_k}$, $\eta_k = \nu_k - D_m + 1$, valid under the $\Sigma \sim \mathrm{IW}(\Psi, \nu)$ convention that \texttt{niw.py} demonstrably uses (Step 2, matching Murphy 2012 eq.~4.222); (c) $w_k(s^*) \propto \mathbb{E}[\pi_k]\,\mathrm{St}_s(s^* \mid k)$ with $p_k(c \mid s^*) = \mathrm{St}_c(c \mid k)$ requiring no correction \emph{because and only because} the per-component predictive factorizes (Step 3); (d) $\mathbb{E}[c \mid s^*] = \sum_k w_k m_{k,c}$ and $\mathrm{Cov}[c \mid s^*] = \sum_k w_k(C_k + m_{k,c}m_{k,c}^{\!\top}) - \bar m \bar m^{\!\top}$ (Step 4); (e) $C_k = \tfrac{\eta_{k,c}}{\eta_{k,c}-2} S_{k,c}$ for $\eta_{k,c} > 2$ (Step 5). The code (\texttt{niw\_predictive}, \texttt{student\_logpdf}, \texttt{conditional\_color\_moments} in \texttt{predictive.py}) implements these formulas exactly, including the $\eta_c > 2$ runtime guard. This establishes Eq.~\eqref{eq:condmoments} exactly as stated in \S\ref{sec:inference}.

\clearpage
\section{Rendering-control decomposition, per scene}
\label{app:rendercontrols}

Table~\ref{tab:rendercontrols} gives the per-scene detail behind the controlled
decomposition that closes \S\ref{sec:raster}. All arms are exported and rendered under the
protocol of Table~\ref{tab:raster} (\texttt{gsplat}, white background, $25$ strictly
held-out views sampled from frame $40$ onward) unless a column varies exactly one element
of it. The fixed-$K$ arm is a symmetric Dirichlet mixture with $e_0 = 1/K$ at $K = \Khat$,
the capacity DP-Splat selects at $\alpha{=}100$ (the $\Khat$ column of
Table~\ref{tab:raster}), fit over three seeds; entries are seed means $\pm$ standard
deviations. The capacity columns rerun that arm at $K$ equal to the stated fraction of
$\Khat$, seed $0$, with the realized live-component count in parentheses; it falls below
the nominal capacity at the larger arms because the fit empties part of what it is given.
The last three columns act on the canonical DP-Splat $\alpha{=}100$ exports of
Table~\ref{tab:raster} with the fits untouched: $\Delta_{\mathrm{soft}}$ is the PSNR change
from replacing binary opacity with the soft count-based map
$o_k = \log(1+N_k)/\log(1+\max_j N_j)$ on the unchanged live mask, and
$\Delta_{s=1.0}$ / $\Delta_{s=2.0}$ are the changes from replacing the inherited scale
modifier $1.41$ with $1.0$ / $2.0$. VBGS is not re-exported under any alternative map,
since $1.41$ and binary opacity are its own protocol's values. Data sources:
\texttt{r7\_render\_dir.json}, \texttt{r7\_render\_cap.json},
\texttt{r7\_render\_dp-soft.json}, and \texttt{r7\_render\_scale\{1.0,1.41,2.0\}.json} in
the experiment records, with the DP-Splat and VBGS reference rows read from
\texttt{r3\_render\_metrics\_v2.json}.

\begin{table}[h]
\centering
\footnotesize
\setlength{\tabcolsep}{3pt}
\caption{Rendering controls on all eight NeRF-synthetic scenes (PSNR in dB; higher is
better for PSNR, and positive is better for the $\Delta$ columns). Left block: fixed-$K$
Dirichlet at the matched capacity $K = \Khat$ (three seeds, mean $\pm$ sd) against the
DP-Splat $\alpha{=}100$ and VBGS references of Table~\ref{tab:raster}. Middle block: the
fixed-$K$ capacity sweep, seed $0$, realized live count in parentheses. Right block:
export-map deltas on the canonical DP-Splat exports. Sources:
\texttt{r7\_render\_dir.json}, \texttt{r7\_render\_cap.json},
\texttt{r7\_render\_dp-soft.json}, \texttt{r7\_render\_scale\{1.0,1.41,2.0\}.json},
\texttt{r3\_render\_metrics\_v2.json}.}
\label{tab:rendercontrols}
\begin{tabular}{lrcrrccccrrr}
\toprule
& & \multicolumn{3}{c}{Matched capacity} & \multicolumn{4}{c}{Capacity sweep, fixed-$K$ (seed 0)}
& \multicolumn{3}{c}{Export deltas (DP)} \\
\cmidrule(lr){3-5} \cmidrule(lr){6-9} \cmidrule(lr){10-12}
Scene & $\Khat$ & fixed-$K$ Dir. & DP & VBGS & $\Khat/4$ & $\Khat/2$ & $\Khat$ & $2\Khat$
& $\Delta_{\mathrm{soft}}$ & $\Delta_{s=1.0}$ & $\Delta_{s=2.0}$ \\
\midrule
\emph{chair}     & 230 & $14.92 \pm 1.55$ & 13.24 & 20.56 & 16.42 (56) & 16.58 (111) & 14.07 (198) & 14.17 (281) & $+4.17$ & $+2.31$ & $-2.44$ \\
\emph{drums}     & 300 & $12.05 \pm 0.30$ & 12.42 & 16.52 & 12.20 (75) & 12.57 (149) & 12.29 (283) & 11.46 (447) & $+1.12$ & $+1.37$ & $-1.40$ \\
\emph{ficus}     & 322 & $13.82 \pm 0.49$ & 13.81 & 17.50 & 14.48 (80) & 14.20 (161) & 13.28 (308) & 13.04 (532) & $+0.78$ & $+2.06$ & $-2.30$ \\
\emph{hotdog}    & 261 & $14.91 \pm 0.14$ & 14.80 & 19.27 & 15.61 (65) & 15.80 (127) & 14.84 (251) & 14.12 (384) & $+1.81$ & $+2.59$ & $-2.74$ \\
\emph{lego}      & 260 & $12.84 \pm 0.73$ & 11.91 & 17.15 & 12.91 (65) & 13.10 (128) & 12.09 (234) & 12.12 (378) & $+2.24$ & $+2.41$ & $-2.46$ \\
\emph{materials} & 315 & $11.83 \pm 0.52$ & 12.07 & 15.47 & 12.69 (79) & 12.61 (158) & 11.87 (311) & 11.89 (493) & $+0.85$ & $+1.88$ & $-2.14$ \\
\emph{mic}       & 164 & $11.23 \pm 0.63$ & 10.56 & 19.60 & 10.88 (41) & 11.74 (77)  & 11.53 (143) & 10.36 (215) & $+1.81$ & $+2.73$ & $-2.29$ \\
\emph{ship}      & 330 & $10.50 \pm 0.24$ & 10.48 & 14.48 & 10.42 (82) & 10.79 (164) & 10.23 (316) & 10.11 (549) & $+0.90$ & $+1.38$ & $-1.67$ \\
\midrule
mean             & 273 & $12.76$          & 12.41 & 17.57 & --- & --- & --- & --- & $+1.71$ & $+2.09$ & $-2.18$ \\
\bottomrule
\end{tabular}
\end{table}

\end{document}